\documentclass[10pt,twocolumn,letterpaper]{article}

\usepackage[pagebackref=true,breaklinks=true,letterpaper=true,colorlinks,bookmarks=false]{hyperref}

\usepackage{times}  % DO NOT CHANGE THIS
\usepackage{helvet}  % DO NOT CHANGE THIS
\usepackage{courier}  % DO NOT CHANGE THIS
\usepackage{amssymb}
\usepackage{amsmath}
\usepackage{multirow}
\usepackage{float}
\usepackage{graphicx}
\usepackage{diagbox}
\usepackage{caption}
\usepackage{algorithm}
\usepackage{color}
\usepackage{algorithmic}
\usepackage{bbding}
\usepackage{booktabs}
\usepackage{gensymb}
\usepackage{makecell}
\usepackage{amsbsy, amsfonts}

\usepackage{bm}
\usepackage{graphicx}
\usepackage{multirow}
\usepackage{subfigure}
\usepackage{enumitem}

\usepackage{newfloat}
\usepackage{listings}
\usepackage{wrapfig}
\usepackage{amssymb,amsmath,amsthm}

\newtheorem{theorem}{Theorem}
\newtheorem{lemma}{Lemma}

\usepackage{amsfonts}
\usepackage{bm}
\usepackage{bbm}
\usepackage[T1]{fontenc}

\def\eqref#1{equation~\ref{#1}}
\def\1{\bm{1}}

\DeclareMathAlphabet{\mathsfit}{\encodingdefault}{\sfdefault}{m}{sl}
\SetMathAlphabet{\mathsfit}{bold}{\encodingdefault}{\sfdefault}{bx}{n}

\usepackage{pifont}
\usepackage[perpage,symbol*]{footmisc}
\DefineFNsymbols{circled}{{\ding{192}}{\ding{193}}{\ding{194}}
{\ding{195}}{\ding{196}}{\ding{197}}{\ding{198}}{\ding{199}}{\ding{200}}{\ding{201}}}
\setfnsymbol{circled}

\newcommand\myfootnotestyle[1]{\ifcase#1 \or \ding{182}\or \ding{183}\or
\ding{184}\or \ding{185}\or \ding{186}\or \ding{187}%
\or \ding{188}\or \ding{189}\or \ding{190}\or \ding{191}\else *\fi\relax}

\def\confName{CVPR}
\def\confYear{2023}

\usepackage{cvpr}              % To produce the CAMERA-READY version
\def\cvprPaperID{4304} % *** Enter the CVPR Paper ID here

\def\eg{\emph{e.g.}}
\def\ie{\emph{i.e.}}

\def\wrt{\emph{w.r.t.}}

\begin{document}

\title{Exploring the Relationship between Architectural Design and\\Adversarially Robust Generalization}
\author{Aishan Liu\textsuperscript{1}, Shiyu Tang\textsuperscript{1}, Siyuan Liang\textsuperscript{2}, Ruihao Gong\textsuperscript{1,6}, Boxi Wu\textsuperscript{3}, \\Xianglong Liu\textsuperscript{1,4,5}\textsuperscript{\thanks{The first three authors contribute equally. Corresponding author: Xianglong Liu, xlliu@buaa.edu.cn.}}, Dacheng Tao\textsuperscript{7}\\
\fontsize{11.0pt}{\baselineskip}
\selectfont 
\textsuperscript{1}Beihang University, \textsuperscript{2}Chinese Academy of Sciences, \textsuperscript{3}Zhejiang University,\\
\textsuperscript{4}Zhongguancun Laboratory, \textsuperscript{5}Hefei Comprehensive National Science Center, \\
\textsuperscript{6}SenseTime,
\textsuperscript{7}JD Explore Academy\\}

\maketitle

\begin{abstract}
Adversarial training has been demonstrated to be one of the most effective remedies for defending adversarial examples, yet it often suffers from the huge robustness generalization gap on unseen testing adversaries, deemed as the adversarially robust generalization problem. Despite the preliminary understandings devoted to adversarially robust generalization, little is known from the architectural perspective. To bridge the gap, this paper for the first time systematically investigated the relationship between adversarially robust generalization and architectural design. In particular, we comprehensively evaluated 20 most representative adversarially trained architectures on ImageNette and CIFAR-10 datasets towards multiple $\ell_p$-norm adversarial attacks. Based on the extensive experiments, we found that, under aligned settings, Vision Transformers (e.g., PVT, CoAtNet) often yield better adversarially robust generalization while CNNs tend to overfit on specific attacks and fail to generalize on multiple adversaries. To better understand the nature behind it, we conduct theoretical analysis via the lens of Rademacher complexity. We revealed the fact that the higher weight sparsity contributes significantly towards the better adversarially robust generalization of Transformers, which can be often achieved by the specially-designed attention blocks. We hope our paper could help to better understand the mechanism for designing robust DNNs. Our model weights can be found at \url{http://robust.art}.

%Thus, this paper tries to bridge the gap by systematically examining the most representative architectures (\eg, Vision Transformers and CNNs). In particular, we first comprehensively evaluated \emph{20} adversarially trained architectures on ImageNette and CIFAR-10 datasets towards several adversaries (multiple $\ell_p$-norm adversarial attacks), and found that Vision Transformers (\eg, PVT, CoAtNet) often yield better adversarially robust generalization. To further understand what architectural ingredients favor adversarially robust generalization, we delve into several key building blocks and revealed the fact via the lens of Rademacher complexity that the higher weight sparsity contributes significantly towards the better adversarially robust generalization of Vision Transformers, which can be often achieved by attention layers. Our extensive studies discovered the close relationship between architectural design and adversarially robust generalization, and instantiated several important insights. We hope our findings could help to better understand the mechanism towards designing robust deep learning architectures.\footnote{Our codes can be found in Supplementary Material.}
\end{abstract}

\section{Introduction}

\begin{figure}[tb]
\centering
\includegraphics[scale=0.33]{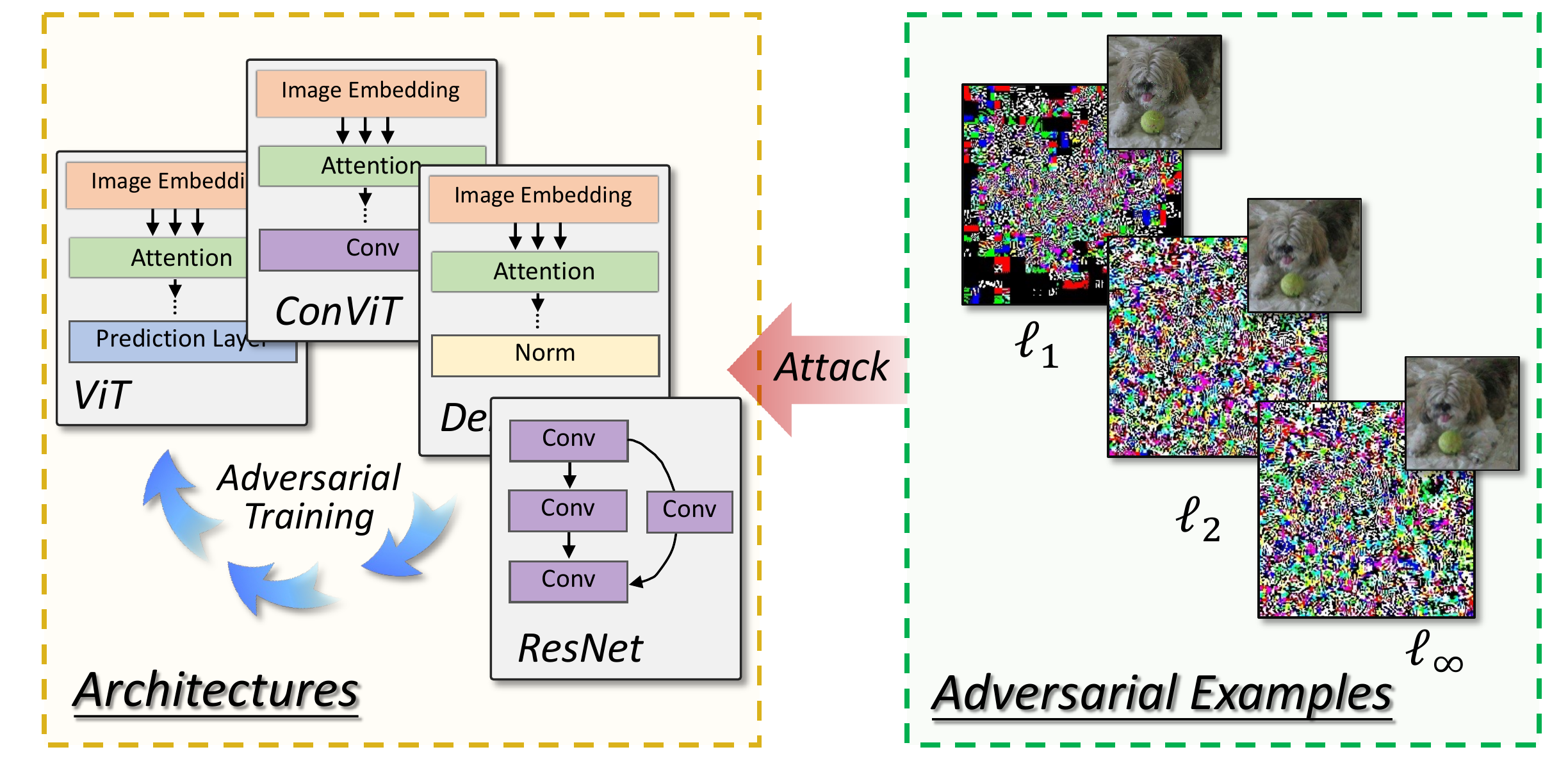}
\caption{This paper aims to better understand adversarially robust generalizations from the architectural design perspective. We comprehensively evaluated 20 adversarially trained architectures on several datasets toward multiple $\ell_p$ adversaries and found that Transformers often yield better adversarially robust generalization. We also undertake theoretical analysis through the Rademacher complexity lens to further comprehend this.}
\label{motivation}
\vspace{-0.2in}
\end{figure}

DNNs have achieved remarkable performance across a wide variety of applications \cite{krizhevsky2012imagenet, he2016deep,zhao2023temporal,ma2022regionwise}, yet they are vulnerable to adversarial examples (malicious inputs that could fool DNNs by adding human-imperceptible perturbations \cite{szegedy2013intriguing, goodfellow2014explaining, liu2020bias, liu2023xadv}). Numerous defensive strategies have been offered to combat this issue. The most effective and promising of these approaches is adversarial training~\cite{tramer2018ensemble, madry2018towards}, which improves adversarial robustness by adding adversarial examples into training data.

However, adversarial training tends to exhibit a huge generalization gap between training and testing data \cite{tsipras2018robustness, su2018robustness, zhang2018limitations}. Unlike models trained on clean data that usually generalize well on clean testing data, differences between training and testing robustness for adversarially-trained models tend to be large. Moreover, the robustness of adversarial training fails to generalize to unseen adversaries during testing \cite{kurakin2016adversarial, laidlaw2020perceptual}. The aforementioned \emph{adversarially robust generalization problem} puts a barrier in the way of the potential applications of adversarial training in practice, where attacks are often drawn from multiple unknown attacks, therefore attracting intensive research interests. Prior works have studied the characteristic of adversarially robust generalization and proposed many approaches for mitigation (\eg, unlabeled data \cite{schmidt2018adversarially, carmon2019unlabeled}, robust local features \cite{song2019robust}, generative adversarial training \cite{poursaeed2021robustness}), while little is known about adversarially robust generalization in terms of architectural design. Model architecture reflects the inherent nature of model robustness \cite{tang2021robustart}. To better comprehend and create robust DNNs, it is important to carefully explore the links between architectural design and robust generalization.

In this paper, we present the first comprehensive investigation on adversarially robust generalization \wrt{} architectural design. In particular, we first provide an extensive evaluation on \emph{20} adversarially trained architectures, including the prevailing Vision Transformers (\eg, ViT, PVT, Swin Transformer) as well as some representative CNNs (\eg, ResNet, VGG) for ImageNette and CIFAR-10 datasets towards multiple adversaries ($\ell_1$, $\ell_2$, $\ell_{\infty}$ adversarial attacks). Based on our experiments, we found that, under aligned setting, Vision Transformers (\eg, PVT, CoAtNet) often yield better adversarially robust generalization compared to CNNs; yet CNNs often tend to overfit on specific attacks and fail to generalize on multiple adversaries. To better understand the mechanism, we then conduct theoretical analysis via the lens of Rademacher complexity. We theoretically revealed that higher weight sparsity contributes significantly to the better adversarially robust generalization of Transformers, which can often be achieved by the specially-designed architectural ingredient attention blocks. At last, we also provided more detailed analyses to better understand generalization (\eg, generalization on common corruptions, larger dataset ImageNet) and discussed the potential directions for designing robust architectures. Our \textbf{main contributions} can be summarized as:

\begin{itemize}
    \item We, for the first time, systematically studied 20 adversarially-trained architectures against multiple attacks and revealed the close relationship between architectural design and robust generalization.
    \item We theoretically revealed that higher weight sparsity contributes to the better adversarially robust generalization of Transformers, which can often be achieved by attention blocks.
    \item We provide more detailed analyses of the generalizability from several viewpoints and discuss potential pathways that may improve architecture robustness.
\end{itemize}

We open-sourced the weights of our adversarially-trained model zoo. Together with our analysis, we hope this paper could help researchers build more robust DNN architectures in the future.
\section{Related Work}

\textbf{Adversarial Attacks.} Adversarial examples are intentionally designed inputs that are imperceptible to human vision but intended to mislead deep learning models~\cite{szegedy2013intriguing,goodfellow2014explaining}, which is denoted as
\begin{equation}
f_{\bm{\theta}}(\bm{x}_{i}+\bm{\delta}_{i}) \neq \bm{y}_{i}, \  \operatorname{ s.t. }\ \bm{\delta}_i \in \mathbb{B}(\epsilon),
\end{equation}
where $\bm{\delta}_i$ is the small perturbation bounded by budgets $\epsilon$ under a certain distance metric $\mathbb{B}$ (usually $\ell_p$-norm) and $\bm{y}_{i}$ is the ground-truth label for the image. A long line of work has been proposed to attack deep learning models \cite{goodfellow2014explaining,kurakin2018adversarial,Liu2019Perceptual,Liu2020Spatiotemporal,wei2022boosting,wei2022towards}, which could be roughly split into white-box and black-box attacks based on how they got access to the victim model.

\textbf{Adversarial Training.} Among the proposed defense methods, adversarial training \cite{goodfellow2014explaining, madry2018towards, wang2019convergence, zhang2020attacks, ding2019mma, shafahi2019adversarial, liu2021ANP, wong2019fast} has been demonstrated to be the most effective one. This technique improves the robustness of DNNs by injecting adversarial examples during training, which can be viewed as a min-max optimization. Formally, given a DNN $f_{\bm{\theta}}$, input image $\bm{x}$ and ground truth label $y$ sampled from data distribution $\mathbb{D}$, the standard adversarial training (SAT) is to minimize the adversarial expected risk as 

\begin{equation}
\hat{R}(f_{\bm{\theta}})=\ \mathbb{E}_{(\bm{x}, y) \sim \mathbb{D}}\left[\max _{\bm{\delta} \in \mathbb{B}(\epsilon)} \mathcal{L}\left(f_{\bm{\theta}}, \bm{x}+\bm{\delta}, \bm{y}\right)\right],
\end{equation}

\noindent where $\mathcal{L}(\cdot)$ is the loss function. %Since we cannot access the true adversarial data distribution, we usually use the empirical distribution to approximate the true distribution. For any data $(\bm{x}_{i}+\bm{\delta}_{i}, \bm{y_{i}})$ with $1/n$ probability, the adversarial empirical risk is defined as
%\begin{equation}
%\hat{R}_{n}(f_{\bm{\theta}})= \frac{1}{n} \sum_{i=1}^{n}\left[\max _{\bm{\delta} \in \mathbb{B}(\epsilon)} \mathcal{L}\left(f_{\bm{\theta}}, \bm{x}_{i}+\bm{\delta}_{i}, \bm{y_{i}}\right)\right].
%\end{equation}

Based on SAT, many different types of adversarial training have been suggested to make adversarial robustness even better. For example, Adversarial Logit Pairing (ALP) \cite{kannan2018adversarial} was proposed to minimize the distance of model logits between clean examples and corresponding adversarial examples; \cite{zhang2019theoretically} stated the trade-off between adversarial robustness and standard accuracy and proposed TRADES, which introduces a regularization term to induce a shift of the decision boundaries away from the training data points; \cite{wang2019improving} considered misclassified examples and incorporated a differentiation regularizer into the adversarial loss. 

\textbf{Adversarially Robust Generalization.} For adversarial training, extensive studies have shown that there is a large gap between training and testing robustness for both seen and unseen adversaries. Several studies have been conducted to better understand this adversarially robust generalization gap. For example, \cite{yu2021understanding} aimed to analyze adversarially robust generalizations via bias-variance decomposition, and found the bias-variance characteristic of adversarial training. \cite{stutz2019disentangling} disentangled robustness and generalization by introducing on-manifold adversarial examples. \cite{tramer2019adversarial} demonstrated that some types of adversarial perturbations (different $\ell_p$-norms) are mutually exclusive, and training on one type of perturbation would not generalize well to other types. \cite{huang2021exploring} explored the influence of depth and width on the adversarially robust generalization for WideResNet. Meanwhile, some research aims to improve adversarially robust generalization to adversarial examples or even other unanticipated perturbations. \cite{schmidt2018adversarially, carmon2019unlabeled} confirmed that more unlabeled training data could close the adversarially robust generalization gap. Besides the huge generalization gap on adversaries, adversarial training is also demonstrated to not generalize well towards clean data, and extensive evidence has been demonstrated that adversarial robustness may be inherently at odds with natural accuracy~\cite{tsipras2018robustness,zhang2019theoretically}.

By contrast, this paper takes the first step toward a systematic study and understanding of adversarially robust generalization from an architectural design point of view. 

\section{Empirical Evaluation}

In this section, we first systematically evaluate 20 adversarially-trained architectures on multiple adversaries.

\begin{table*}[tb]
\vspace{-0.1in}
\centering
\caption{Evaluation results on CIFAR-10. We use worst-case robust accuracy to measure robust generalization towards adversarial data (higher the better). Vanilla Acc denotes the clean accuracy of standard training. Results are ranked by the worst-case robust accuracy.}
\label{tab:CifarRobustGen}
\resizebox{1\linewidth}{!}{
\scriptsize
\begin{tabular}{llc|cccccc}
\toprule
                      &                     & \multicolumn{1}{l|}{} & \multicolumn{6}{c}{\textbf{PGD-$\ell_{\infty}$ Adversarial Training}}                                                                                         \\
\textbf{Architecture} & \textbf{Params (M)} & \textbf{Vanilla Acc}  & \textbf{Clean Acc} & \textbf{PGD-$\ell_{\infty}$} & \textbf{AA-$\ell_{\infty}$} & \textbf{PGD-$\ell_{2}$} & \textbf{PGD-$\ell_{1}$} & \textbf{Worst-case Acc} \\ \midrule
PVTv2                 & 12.40               & 88.34                 & 75.99              & 46.48                        & 38.18                       & 35.77                   & 46.14                   & 33.54                   \\
CoAtNet               & 16.99               & 90.73                 & 77.73              & 48.27                        & 39.85                       & 33.80                   & 42.30                   & 32.17                   \\
ViT                   & 9.78                & 86.73                 & 78.76              & 46.02                        & 38.00                       & 30.86                   & 39.27                   & 29.24                   \\
CPVT                  & 9.49                & 90.34                 & 78.57              & 45.02                        & 36.73                       & 30.15                   & 39.22                   & 28.47                   \\
ViTAE                 & 23.18               & 88.24                 & 75.42              & 40.53                        & 33.22                       & 29.67                   & 40.02                   & 28.13                   \\
MLP-Mixer             & 0.68                & 83.43                 & 62.86              & 38.93                        & 31.81                       & 29.27                   & 36.50                   & 27.42                   \\
PoolFormer            & 11.39               & 89.26                 & 73.66              & 46.33                        & 38.93                       & 28.84                   & 34.32                   & 27.36                   \\
CCT                   & 3.76                & 92.27                 & 81.23              & 49.21                        & 40.97                       & 28.29                   & 34.59                   & 26.82                   \\
VGG                   & 14.72               & 94.01                 & 84.30              & 50.87                        & 41.66                       & 26.78                   & 31.48                   & 25.32                   \\
Swin Transformer      & 27.42               & 91.58                 & 80.44              & 48.61                        & 41.31                       & 26.58                   & 30.47                   & 25.04                   \\ %\midrule
LeViT                 & 6.67                & 89.01                 & 77.10              & 47.16                        & 39.87                       & 26.28                   & 29.58                   & 25.04                   \\
MobileViT             & 5.00                & 91.47                 & 77.52              & 49.51                        & 41.50                       & 26.96                   & 29.35                   & 24.41                   \\
BoTNet                & 18.82               & 94.16                 & 80.76              & 51.29                        & 42.95                       & 25.84                   & 27.38                   & 23.15                   \\
WideResNet            & 55.85               & 96.47                 & 89.54              & 55.17                        & 44.13                       & 22.55                   & 23.68                   & 20.88                   \\
DenseNet          & 1.12                    &  94.42                     &  83.23                  &    53.06                          &   44.02                         &  22.55                      &  21.87                       &   19.48                      \\
PreActResNet          & 23.50               & 95.86                 & 87.96              & 54.85                        & 45.81                       & 18.60                   & 16.46                   & 15.11                   \\
CeiT                  & 5.56                & 85.24                 & 71.55              & 36.20                        & 28.02                       & 15.31                   & 16.77                   & 14.35                   \\
ResNet                & 23.52               & 95.60                 & 87.92              & 54.18                        & 45.40                       & 17.52                   & 15.90                   & 14.32                   \\
ResNeXt               & 9.12                & 95.64                 & 87.12              & 51.51                        & 42.66                       & 15.07                   & 13.64                   & 12.18                   \\  
CvT                   & 19.54               & 87.81                 & 73.76              & 41.36                        & 33.67                       & 12.75                   & 9.25                    & 8.76                    \\  %\midrule
\bottomrule
\end{tabular}}
\end{table*}

\subsection{Evaluation protocols}
\label{sec:setting}
We first briefly introduce our evaluation protocols. \emph{More details can be found in Supplementary Material.}

\textbf{Datasets.} We conduct extensive experiments on both CIFAR-10 \cite{krizhevsky2009learning} and ImageNette \cite{imagenette}. CIFAR-10 contains 10 classes with input size 32$\times$32$\times$3; ImageNette is a subset of 10 classes from ImageNet with input size 224$\times$224$\times$3. We also conduct experiments on large-scale dataset ImageNet \cite{krizhevsky2009learning} with 1,000 classes as shown in Section \ref{sec:analysis}.

\textbf{Architectures.} We consider 20 representative model architectures with different categories and designs (\eg, CNN with convolutions, ViT with attentions, hybrids with both attention/convolutions, and newly designed attentions), aiming to find the influential parts. Specifically, we use ViT \cite{dosovitskiy2020image}, MLP-Mixer \cite{tolstikhin2021mlp}, PoolFormer \cite{yu2021metaformer}, Swin Transformer \cite{liu2021swin}, ViTAE \cite{xu2021vitae}, CCT \cite{hassani2021escaping}, MobileViT \cite{mehta2021mobilevit}, CPVT \cite{chu2021conditional}, BoTNet \cite{srinivas2021bottleneck}, CeiT \cite{yuan2021incorporating}, CoAtNet \cite{dai2021coatnet}, CvT \cite{hassani2021escaping}, LeViT \cite{graham2021levit}, PVTv2 \cite{DBLP:journals/corr/abs-2106-13797}), ResNet \cite{he2016deep}, ResNeXt \cite{xie2017aggregated}, WideResNet \cite{zagoruyko2016wide}, PreActResNet \cite{he2016identity}, VGG \cite{simonyan2014very}, and DenseNet \cite{huang2017densely}. \emph{Note that model sizes differ greatly among architectures, and simply aligning model sizes would cause training collapse. Our studies in Section \ref{sec:4_1} also reveal that model sizes have a comparatively small influence on adversarially robust generalizations. Thus, we primarily focus on discussing the influence of architectural design.}

\textbf{Training settings.} For each model architecture, we conduct standard training (vanilla training) and PGD-$\ell_{\infty}$ adversarial training ($\epsilon=8/255$). To avoid the influence of training techniques on adversarially robust generalization, for all models we use the aligned training settings following \cite{bai2021transformers,tang2021robustart}. Specifically, for CIFAR-10 we use AutoAugment \cite{cubuk2018autoaugment} data augmentation, AdamW \cite{loshchilov2018decoupled} optimizer and cosine learning rate with warmup; we set batch size=128 and training epoch=100. For ImageNette we also use AdamW and cosine learning rate with warmup, and set batch size=128 and training epoch=100. Due to the bad convergence of adversarially trained vision transformers using autoaugment on ImageNet \cite{bai2021transformers}, we use the standard data augmentation for ImageNette dataset. To demonstrate that our adversarial-trained models do not suffer from adversarial overfitting \cite{Rice2020Overfitting}, we report the robust training/testing accuracy at every ten epochs in Supplementary Material. \emph{We also conduct comparisons for different architectures using optimal settings (see Supplementary Material).}

\textbf{Evaluation strategy.} To evaluate the adversarially robust generalization, we adopt commonly-used adversarial attacks, including PGD-$\ell_{1}$ \cite{madry2018towards}, PGD-$\ell_{2}$, PGD-$\ell_{\infty}$ and AutoAttack-$\ell_{\infty}$ attacks \cite{croce2020reliable}. For CIFAR-10 we set $\epsilon=8/255, 1.5, 40.0$ for $\ell_{\infty}$, $\ell_{2}$ and $\ell_{1}$ attacks respectively, while for ImageNette we set $\epsilon=8/255, 8.0, 1600.0$ for the three attack norms. As for evaluation metrics, following \cite{carlini2019evaluating}, we use the classification accuracy for clean examples and measure the sample-wise worst-case robust accuracy for adversaries, which is the lower bound for robustness under multiple adversarial attacks. The metric is formulated as

\begin{equation}
W(f_{\bm{\theta}}, \mathbb{A})= \frac{1}{n} \sum_{i=1}^{n}\left\{
\min_{\mathcal{A} \in \mathbb{A}} \bm{1}\left[ f_{\bm{\theta}}\left(\mathcal{A}\left( \bm{x}_{i} \right)\right)=\bm{y}_{i} \right]
\right\},
\end{equation}

\noindent where $(\bm{x}_{i}, \bm{y}_{i})$ is the clean sample pair, $\mathcal{A}$ is attack in the set of multiple adversaries $\mathbb{A}$, and $\bm{1}\left(\cdot\right)$ is the indicator function. For the above metrics, the higher the better.

\subsection{Empirical results}
\label{sec:4_1}

Here, we show the evaluation results of these architectures on CIFAR-10 and ImageNette. Due to space limitation, our main paper only reports the evaluation results of standard trained models and PGD-$\ell_{\infty}$ adversarially trained models for all the architectures on CIFAR-10; the results on ImageNette are shown in Supplementary Material. 

As shown in Table \ref{tab:CifarRobustGen}, for the commonly-adopted $\ell_{\infty}$-norm adversarial attacks, CNNs achieve better robustness than Transformers. However, for $\ell_{1}$ and $\ell_{2}$-norm adversarial attacks, CNNs (especially ResNet and its variants) have much lower robust accuracy than most of the Transformers (\eg, on CIFAR-10, accuracy under PGD-$\ell_{1}$ attack for ResNet and PVTv2 are 15.90 and 46.14). Considering the much more comprehensive worst-case robust accuracy, some representative Transformers (\eg, PVTv2, CoAtNet) achieve the highest robustness and outperform CNNs (\eg, ResNet) by large margins. We conjecture that CNNs might overfit the $\ell_{\infty}$ adversarial attacks during training thus achieving better test accuracy under PGD-$\ell_{\infty}$ while worse accuracy under PGD-$\ell_{1}$ and $\ell_{2}$ adversaries. In general, we could reach several \textbf{observations} as follows.

\ding{182} Architectural design is highly correlated with adversarially robust generalization, resulting in significant influences (\eg, at most 20+\%).

\ding{183} Transformer-based architectures PVTv2 and CoAtNet achieve the best adversarially robust generalization among all architectures. Compared to other ViTs, they contain more attention blocks
rather than convolutional layers (ViTAE, CeiT), more global attention rather than window attention (Swin Transformer), and have a hierarchical structure with diverse sizes, which
are harder to overfit specific attacks during adversarial training.

\ding{184} CNNs (especially ResNet and its variations) tend to overfit trained $\ell_{\infty}$ adversarial attacks, making them less resistant to other attacks and less able to generalize. Besides the worst-case robustness, we also calculate the rank based on each attack
and take the average, where we reach similar observations (\ie, CoAtNet and PVTv2 rank first, and most ViTs outperform CNNs).

\textbf{Other adversarial training methods.} Aside from the conclusions reached on classic PGD-$\ell_{\infty}$ adversarial training, we investigate whether similar conclusions hold up on other variants of adversarial training strategies, where we conduct the same experiments theoretically on CIFAR-10 using TRADES \cite{zhang2019theoretically}. Overall speaking, as shown in Supplementary Material, we could still reach similar observations, demonstrating architectures' consistent influence on adversarially robust generalization.

\textbf{Model size.} Besides the architectural design, model capacity (parameter sizes) also influences robustness \cite{bai2021transformers}. We here choose 3 model pairs that have similar model parameters, \ie, ResNeXt (9.12M) and ViT (9.78M), ResNet (23.52M) and ViTAE (23.18M), DenseNet (1.12M) and MLP-Mixer (0.68M), and compare them in pairs. Table \ref{tab:CifarRobustGen} shows a similar trend under aligned model sizes, namely that transformers frequently have better adversarially robust generalizations than CNNs. Besides, according to Table \ref{tab:CifarRobustGen}, the relationship between model size and adversarially robust generalization is comparatively small, and larger models (\eg, WideResNet) do not necessarily exhibit better generalization than smaller models (\eg, ViT). 

Based on the above results, we could observe that architectural design significantly influences an adversarially robust generalization and that Transformers show superiority over other models. In the next section, we do a theoretical analysis to learn more about how it works.

\section{Theoretical Understanding}
\label{sec:4_2}

Recently, many studies \cite{yin2019rademacher, tu2019theoretical, khim2018adversarial} have theoretically revealed the relationship between adversarial risk and Rademacher complexity, which is a classical metric for measuring generalization error. In this part, we use Rademacher complexity \cite{bartlett2002rademacher} to better understand the above phenomenon and conclusions. 

\subsection{Overall understanding: weight sparsity} 
\begin{figure*}[tb]
\vspace{-0.15in}
\subfigure[ViT]{
\includegraphics[width=0.23\linewidth]{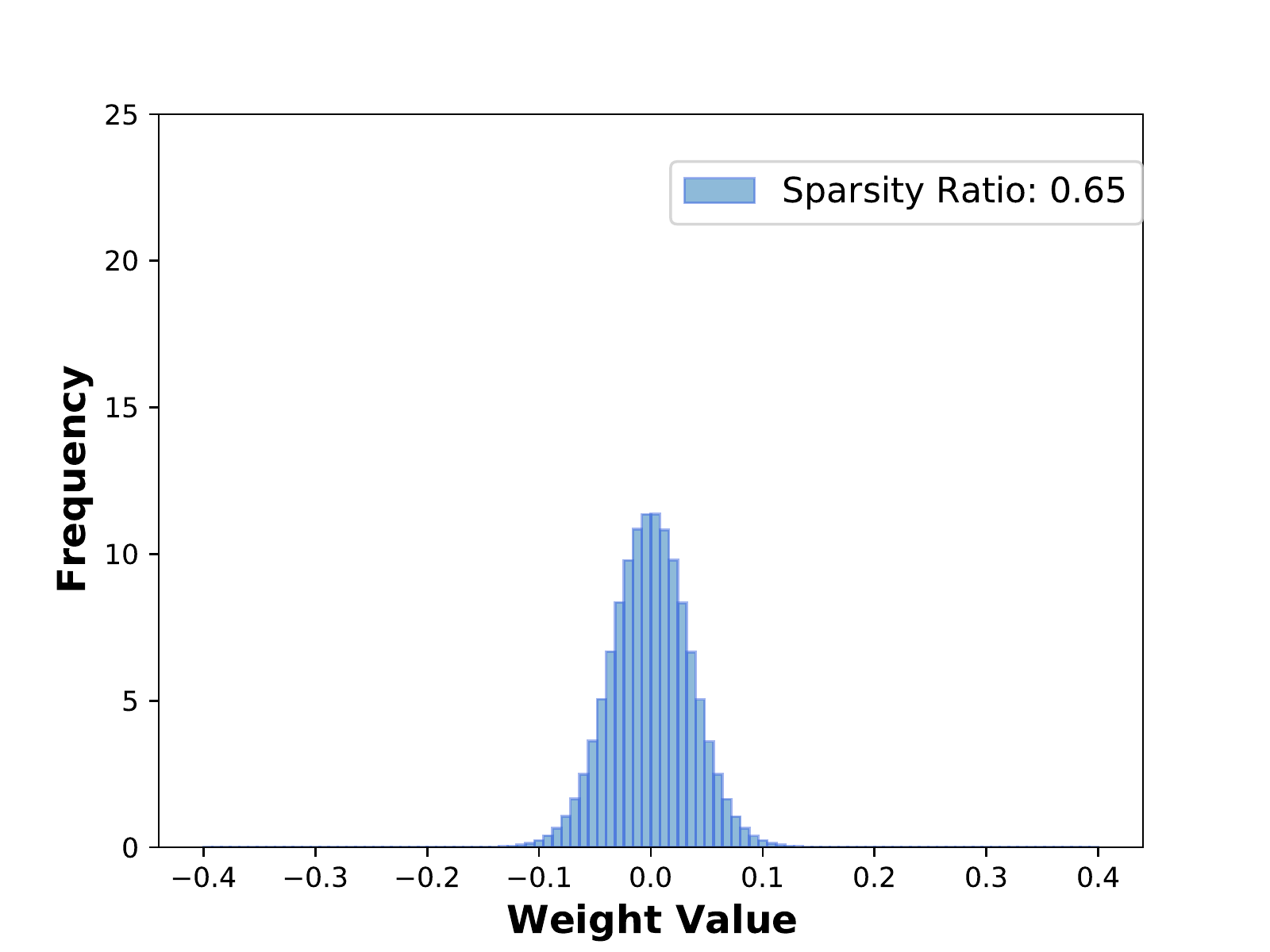}
 \label{fig:sparse-a}
}
\subfigure[PVTv2]{
\includegraphics[width=0.23\linewidth]{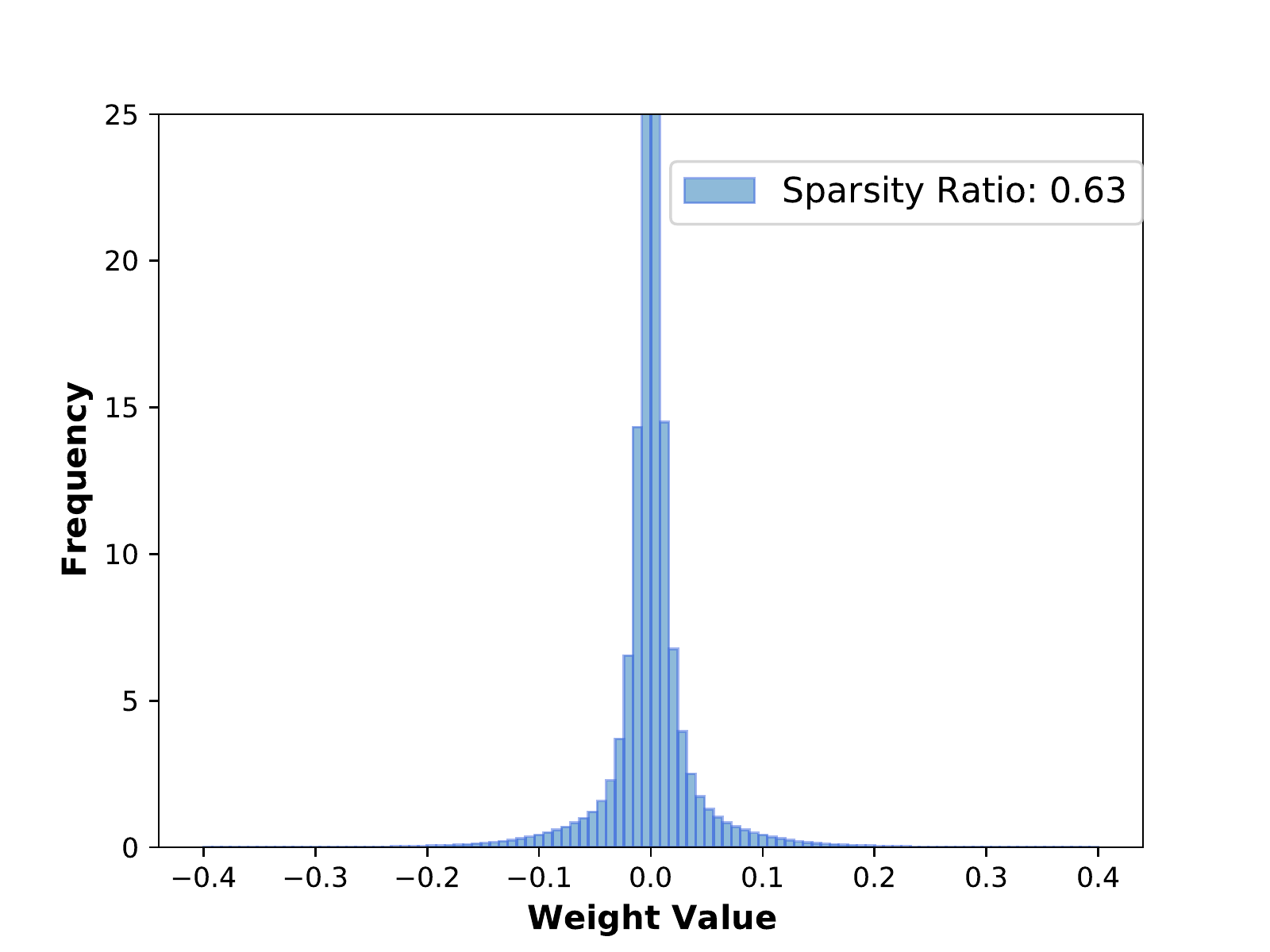}
 \label{fig:sparse-b}
}
\subfigure[CoAtNet]{
\includegraphics[width=0.23\linewidth]{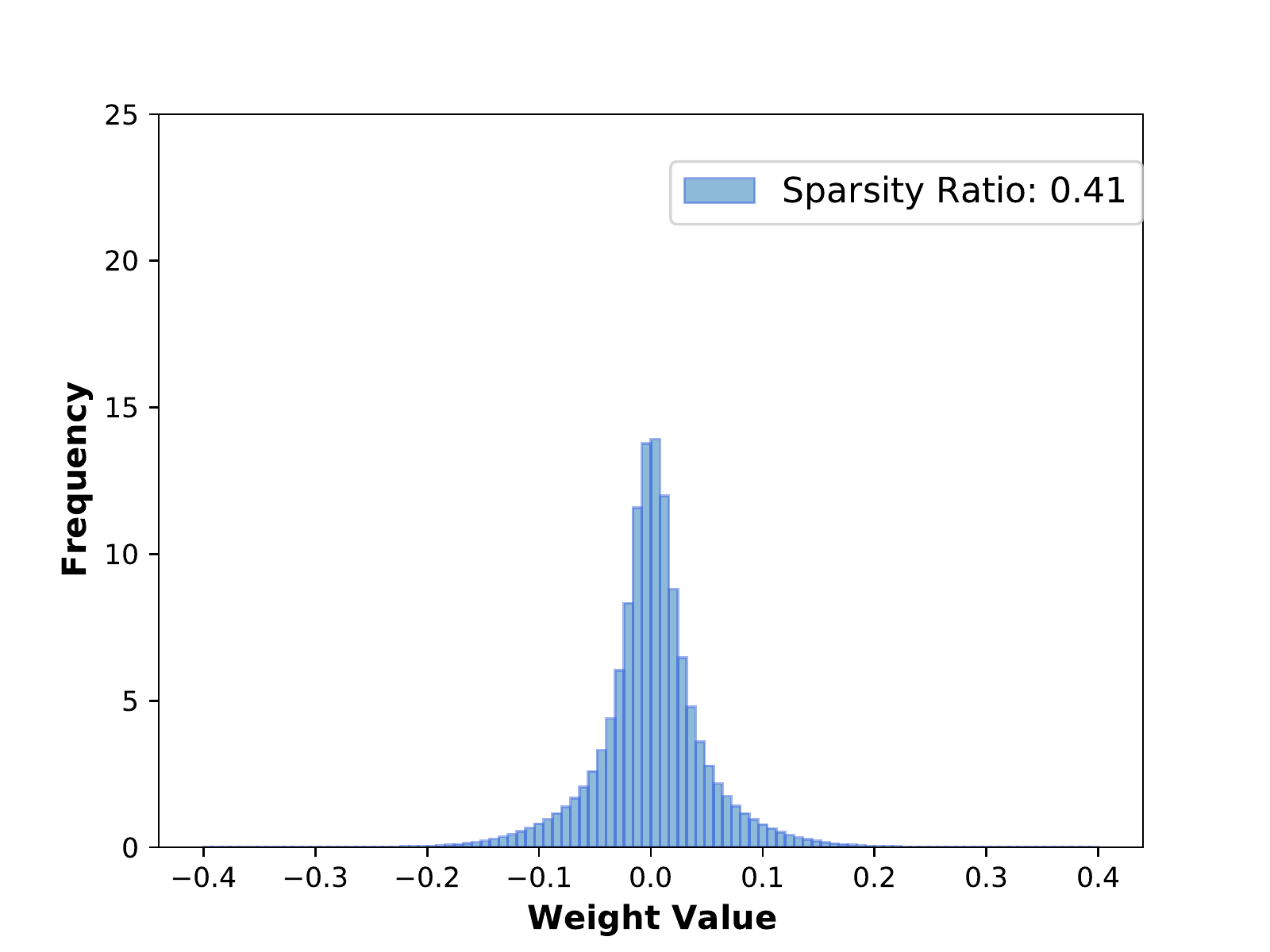}
 \label{fig:sparse-c}
}
\subfigure[CvT]{
\includegraphics[width=0.23\linewidth]{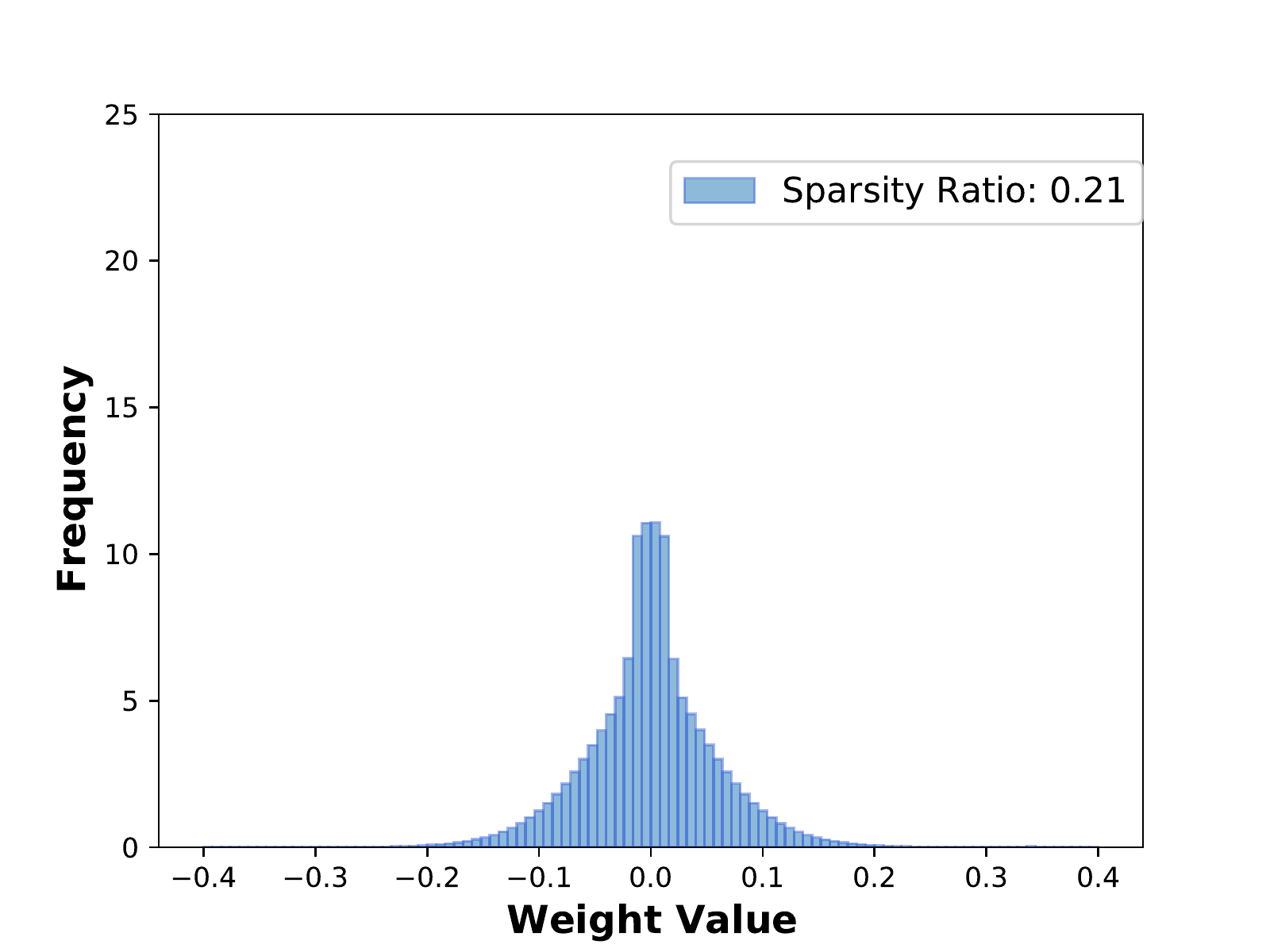}
 \label{fig:sparse-d}
}
\subfigure[VGG]{
\includegraphics[width=0.23\linewidth]{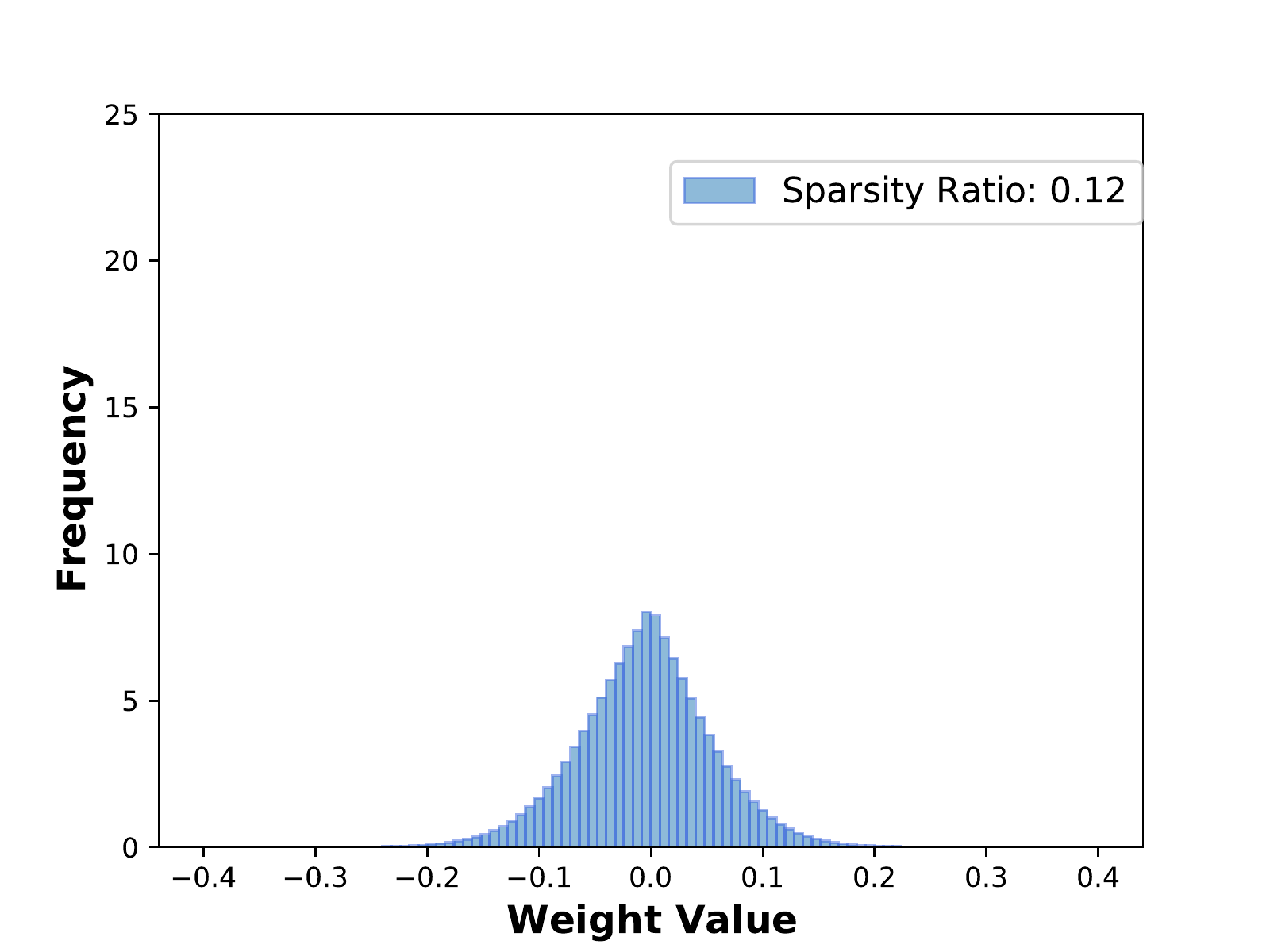}
 \label{fig:sparse-e}
}
\subfigure[DenseNet]{
\hspace{0.02in}
\includegraphics[width=0.23\linewidth]{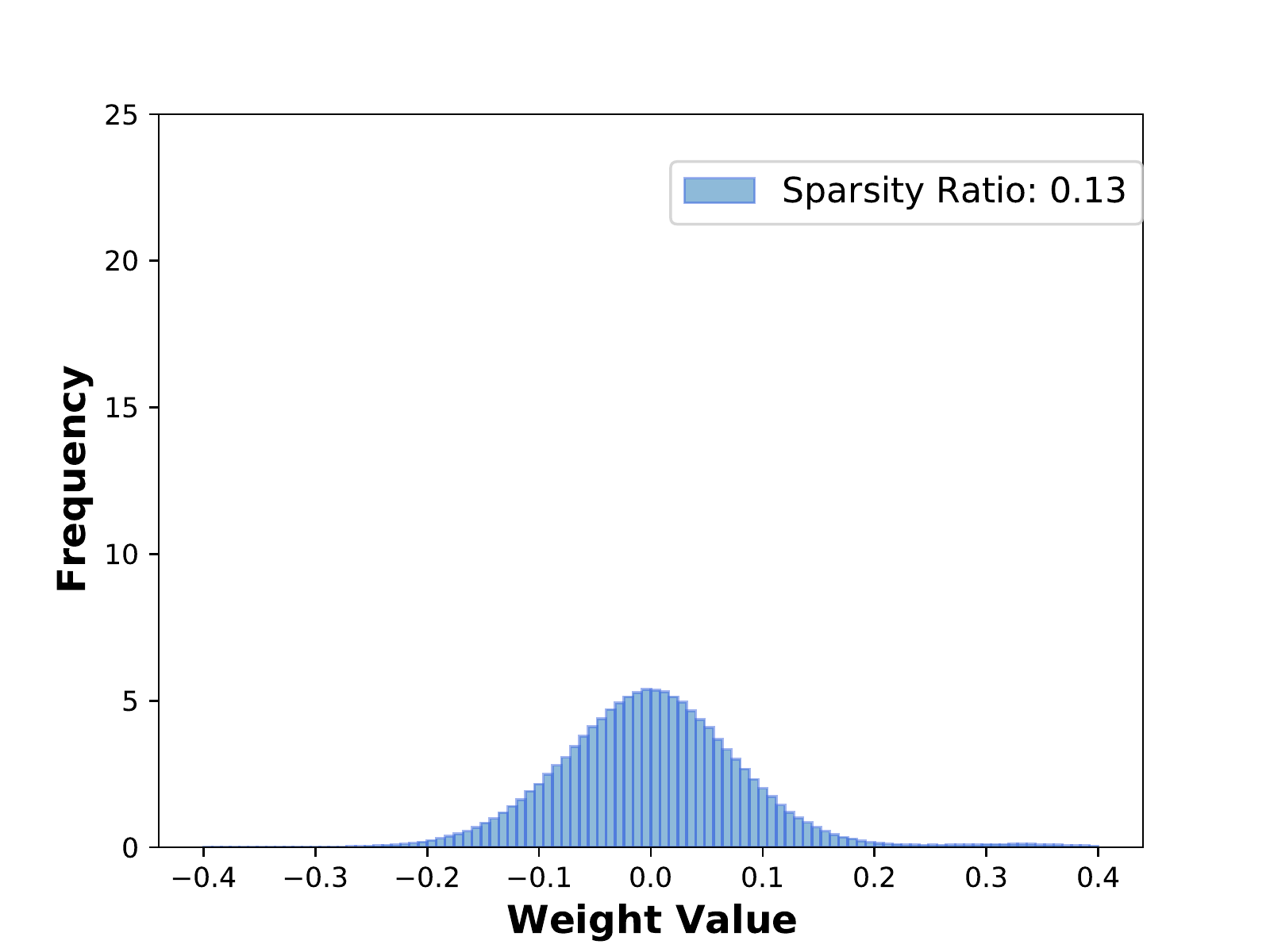}
 \label{fig:sparse-f}
}
\subfigure[ResNet]{
\hspace{0.03in}
\includegraphics[width=0.23\linewidth]{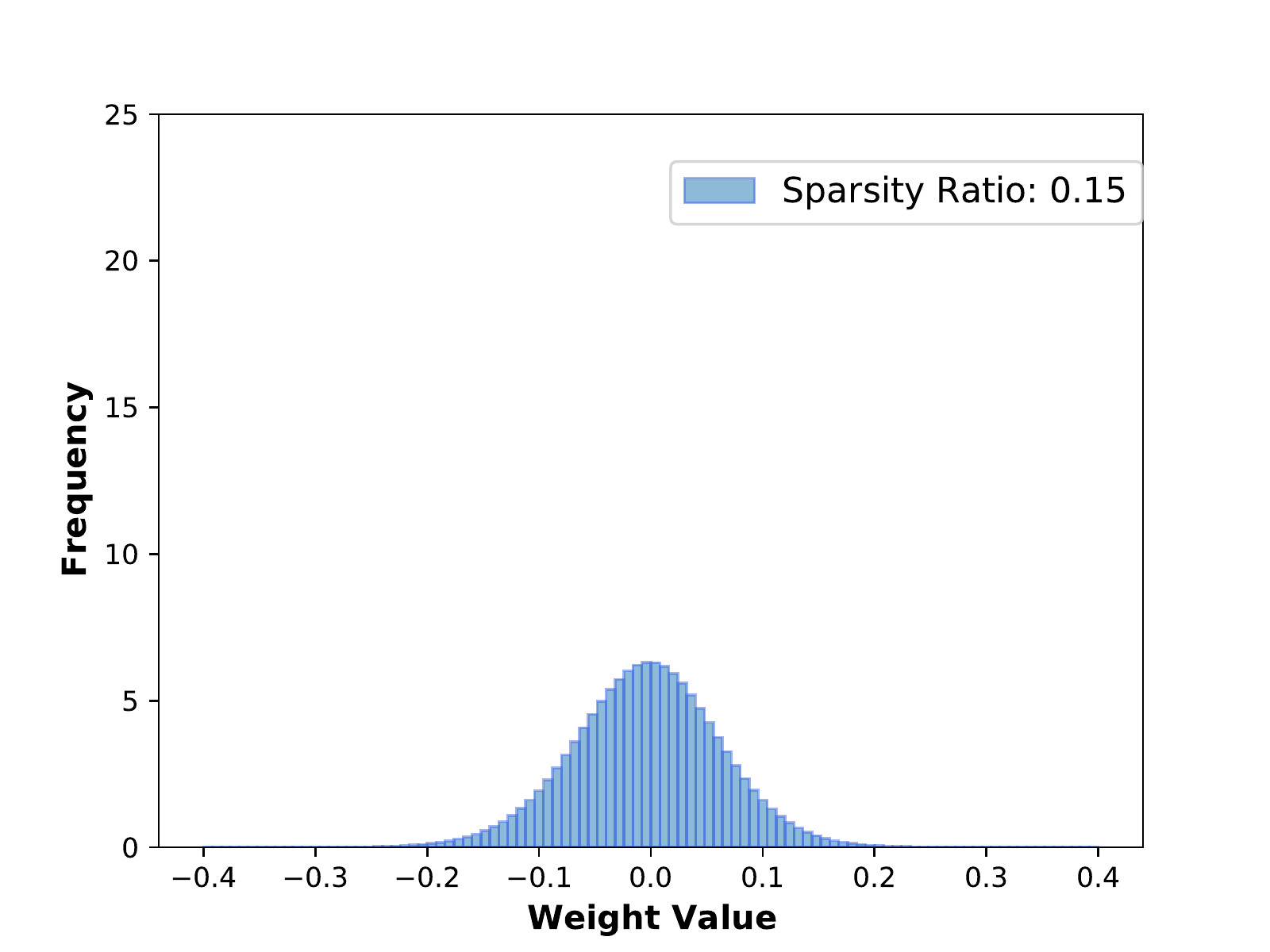}
 \label{fig:sparse-g}
}
\subfigure[WideResNet]{
\hspace{0.02in}
\includegraphics[width=0.23\linewidth]{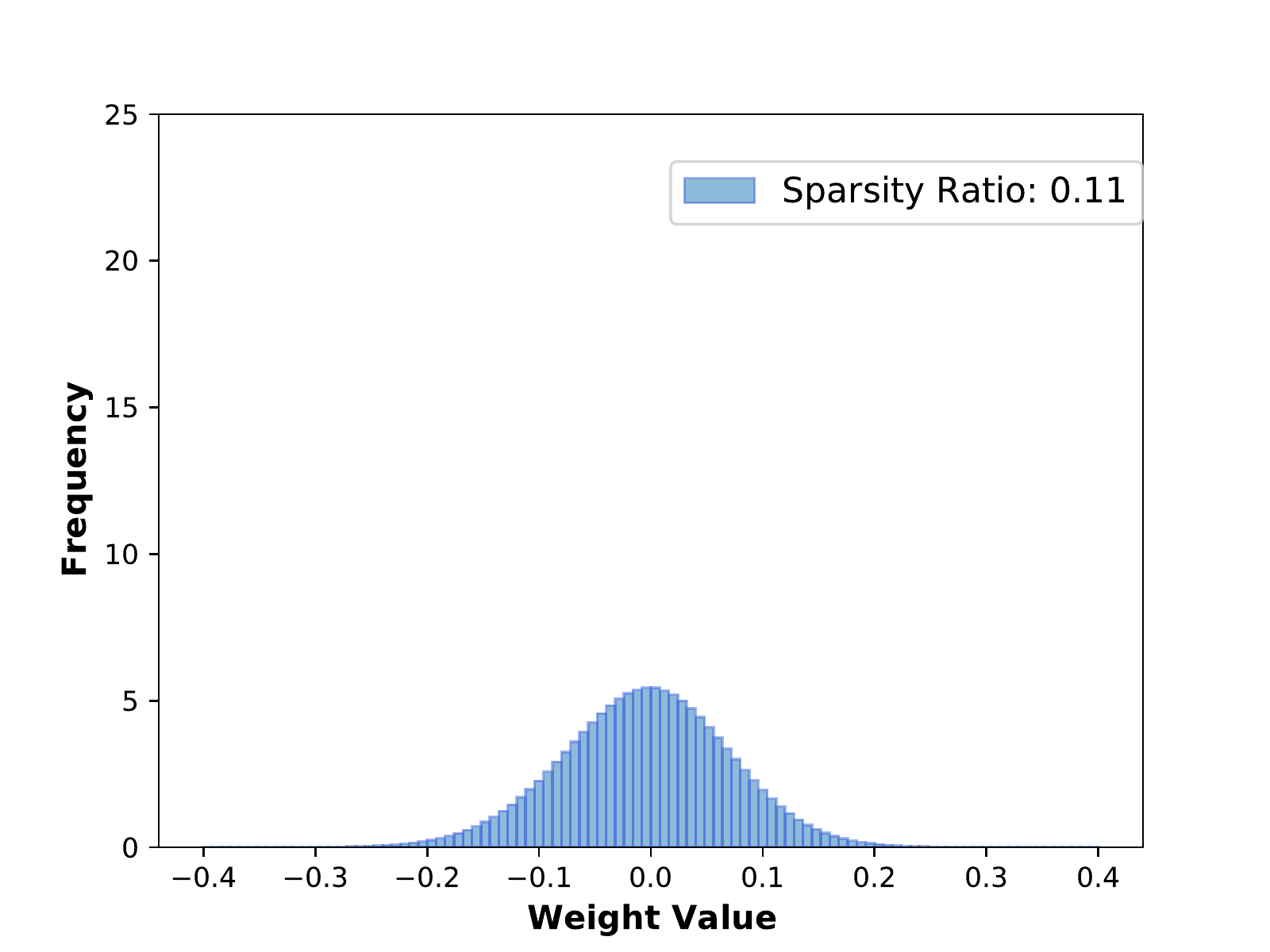}
 \label{fig:sparse-h}
}
\vspace{-0.1in}
\caption{Visualization of weight distribution for adversarially trained models. We calculate the sparsity ratio, which represents the proportion of weights whose absolute value is below a small margin in all weights (we use 0.01 * maximum weight value). Note that the ratio does not indicate the number of zero weights. For example, PVTv2/CoAtNet are narrower and have more zero weights, which are sparser than ViT (flatter) and show better robustness.} 
\label{fig:weightsparse}
\vspace{-0.1in}
\end{figure*}

We first define the Rademacher complexity of the set of all possible evaluation functions $f\in \mathcal{F}$ concerning the training dataset $S$ as follows:
\begin{equation}
R_{\mathcal{S}}(\mathcal{F})= \frac{1}{n} \mathbb{E}_{\bm{\sigma}}\left[\sup_{f \in \mathcal{F}} \sum _{i=1}^{n} \sigma_{i} f(\bm{x}_{i})\right],
\end{equation}

\noindent where the random variable $\sigma_{i} \in [-1, +1]$ with probability 0.5 and $\mathbb{E}_{\bm{\sigma}}$ denotes the expectation. We discuss a linear classifier with binary classification, \ie, the label $\bm{y}\pm 1$. We consider imposing different $\ell_p$ norm on the weight $\theta$ of classifier $f_{\bm{\theta}}$ and define the set of different weight norm classifiers as the function class $\mathcal{F}=\{f_{\bm{\theta}}:\parallel \bm{\theta} \parallel_p \leqslant \Theta\}$. For adversarial training, the function can be written as $\hat{\mathcal{F}}=\{ \min_{\bm{\delta} \in \mathbb{B}(\epsilon)}\bm{y} f_{\bm{\theta}}(\bm{x}+\bm{\delta}): \parallel \bm{\theta} \parallel_p \leqslant \Theta\}$. 

\begin{lemma}
\label{th:th1}
\cite{yin2019rademacher} Suppose that $\frac{1}{p}+\frac{1}{q}=1$, there exists a constant $c\in (0,1)$ such that
\begin{equation}
\label{eq4}
%\begin{aligned}
    \frac{c}{2}(R_{\mathcal{S}}(\mathcal{F})+\epsilon \Theta \frac{d^{1-\frac{1}{p}}}{\sqrt{n}}) \leqslant R_{\mathcal{S}}(\hat{\mathcal{F}}) \leqslant R_{\mathcal{S}}(\mathcal{F})+\epsilon \Theta \frac{d^{1-\frac{1}{p}}}{\sqrt{n}},
%\end{aligned}
\end{equation}
\end{lemma}
\noindent where $d$ denotes the dimension of the input sample, $p$ is the matrix norm of model weights, and $q$ is the norm associated with the input image in Rademacher complexity.

From Lemma~\ref{th:th1}, we can observe that \emph{the Rademacher complexity of adversarial training depends on the classifier $f$, which implies that adversarially robust generalization is influenced by the architecture of the classifier.}

Rademacher complexity relates to the input sample dimension $d$ and the matrix norm $p$ of weight. When $p=1, d^{1-\frac{1}{p}} \rightarrow 1$, the upper and lower bounds of Rademacher complexity are reduced (compared to $p > 1$). More specifically, we can observe that $R_{\mathcal{S}}(\hat{\mathcal{F}})$ has an unavoidable polynomial dimension dependence on $d$ unless the $p=1$. In other words, imposing the $\ell_1$ norm constraints on the weight of the classifier would lead to a sparser weight matrix and reduce the Rademacher complexity of adversarial training. Thus, we visualize the weight distribution by plotting the histogram for multiple adversarially trained architectures on CIFAR-10 in Figure \ref{fig:weightsparse}. From the weight distribution visualization above, we can observe that for PVTv2 and CoAtNet, most weight values are very close to zero, indicating that the two networks have the {\textbf{highest weight sparsity}}. As for the CNNs, ResNet and its variants (\eg, WideResNet, PreActResNet) exhibit bad adversarially robust generalization due to the overfitting on the trained adversarial attacks, and their weight distributions are also \textbf{much denser} than PVTv2 and CoAtNet according to the plots. 

\subsection{Attention contributes to sparseness} 

%As shown in Lemma~\ref{th:th1}, the sparsity of weight would influence adversarially robust generalization. 
%To further investigate the weight sparsity that is affected by which modules of the Transformer network, we consider a toy Transformer. 
Supposing a fundamental block in Transformers $f_{\bm{w}}(\bm{x})=\bm{W}_1\phi(\bm{A}_1\bm{x})$, where $\bm{A}_1=\text{SoftMax}(\bm{Q}\bm{K}^{\top}/\sqrt{d}+\bm{B})\bm{V}$ denotes a Mutil-head Self-Attention layer module, $\bm{W}_{1}$ represents a Multi-Layer Perceptron layer module, and $\phi$ denotes a non-linear activation function, such as ReLU. $\bm{B}$ is the bias term of models. Thus, we extend the derivation from~\cite{yin2019rademacher} that imposing $\ell_1$ constraint in transformer will improve the generalization. In the theory of~\cite{raghunathan2018certified}, the model's generalization cannot be directly optimized, so the upper bound of the model's generalization is found step by step through SDP relaxation, and the generalization is improved by minimizing the upper bound of the model. We can find a generalization upper bound for the transformer model by SDP relaxation and optimize this upper bound to get Theorem~\ref{th:th2} as follows:

\begin{figure*}[tb]
\vspace{-0.1in}
\subfigure[All Weights]{
\includegraphics[width=0.23\linewidth]{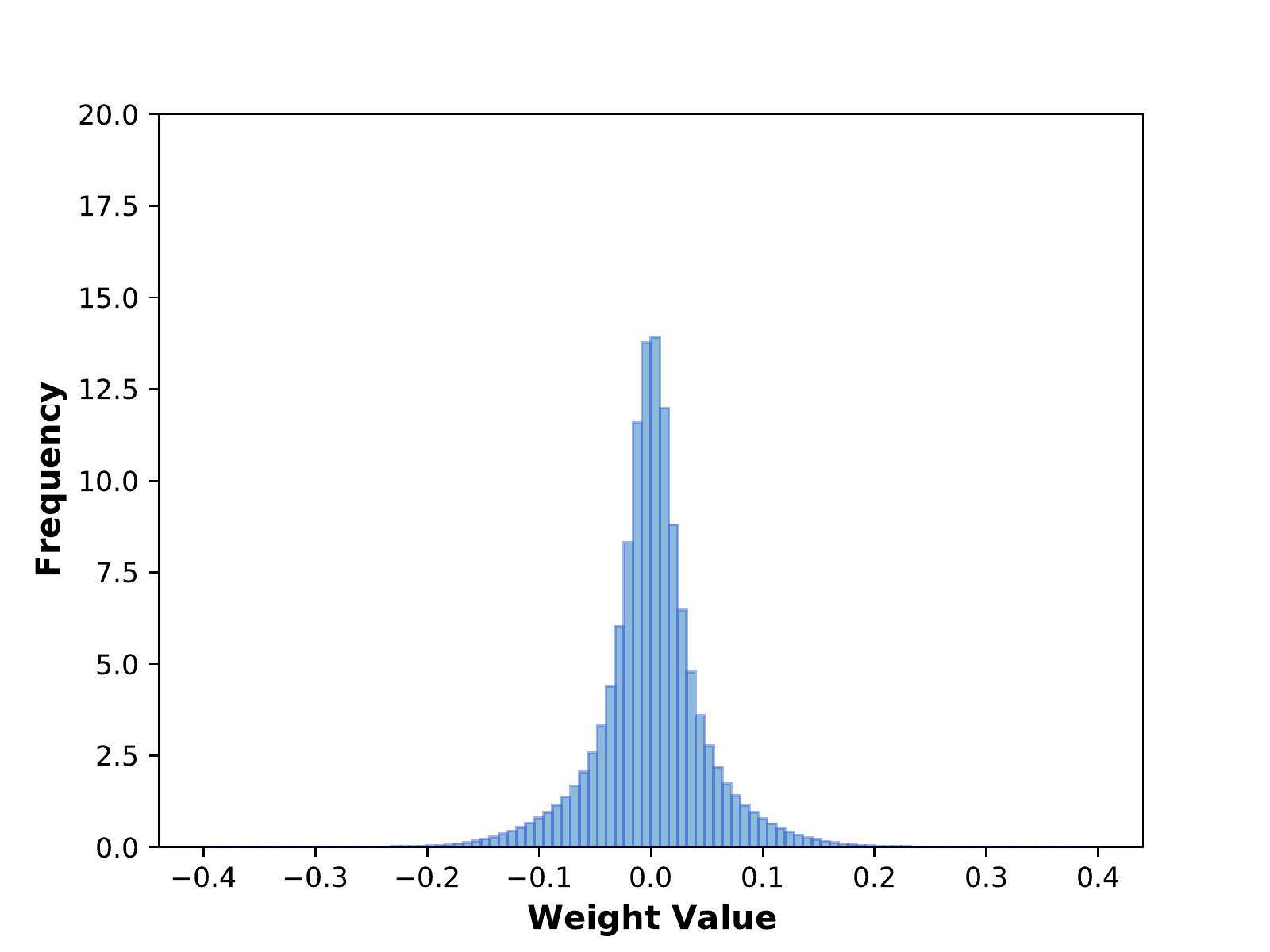}
 \label{fig:coatnetsparse-a}
}
\subfigure[Conv Weights]{
\includegraphics[width=0.23\linewidth]{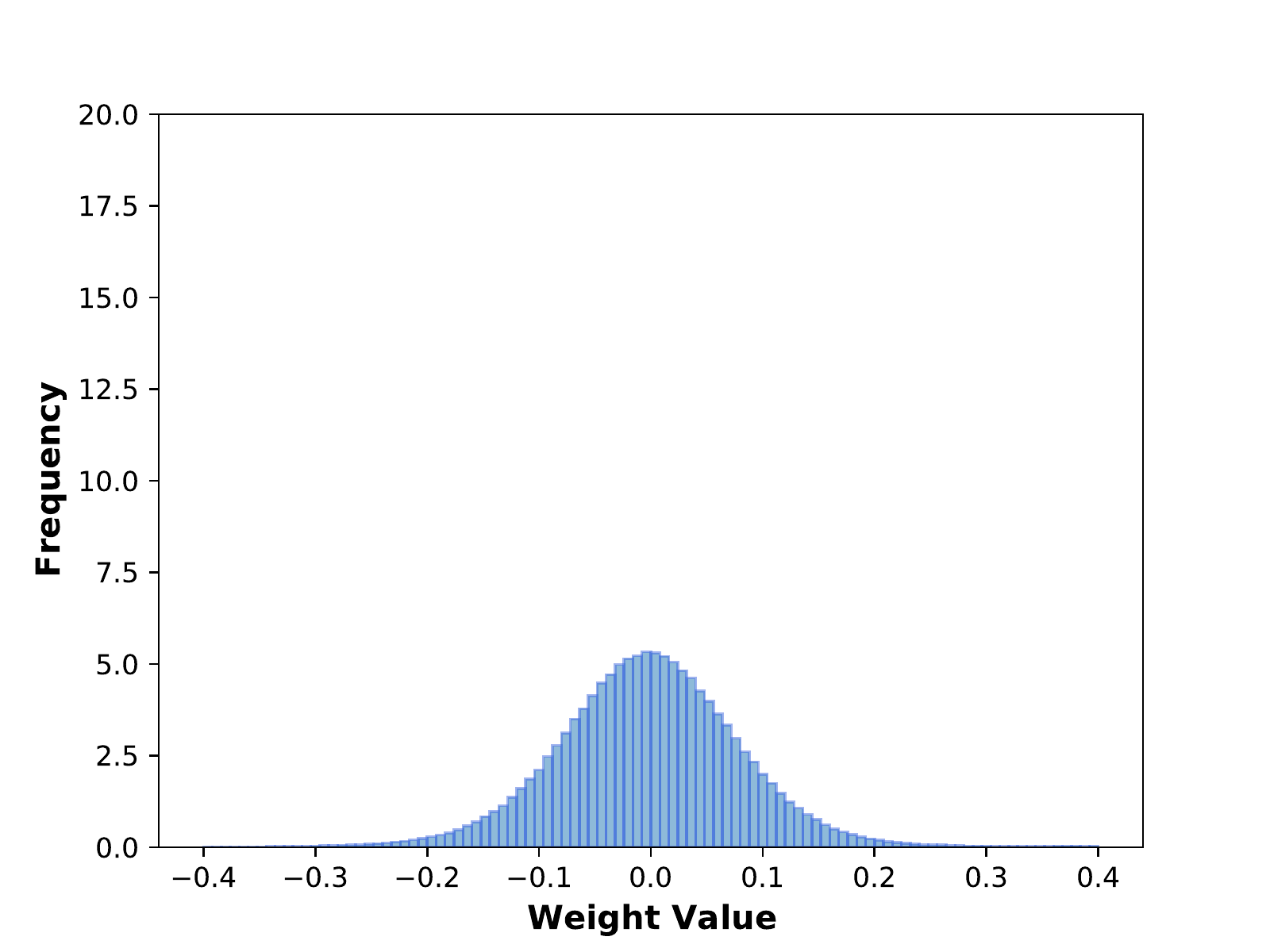}
 \label{fig:coatnetsparse-b}
}
\subfigure[QKV Weights]{
\includegraphics[width=0.23\linewidth]{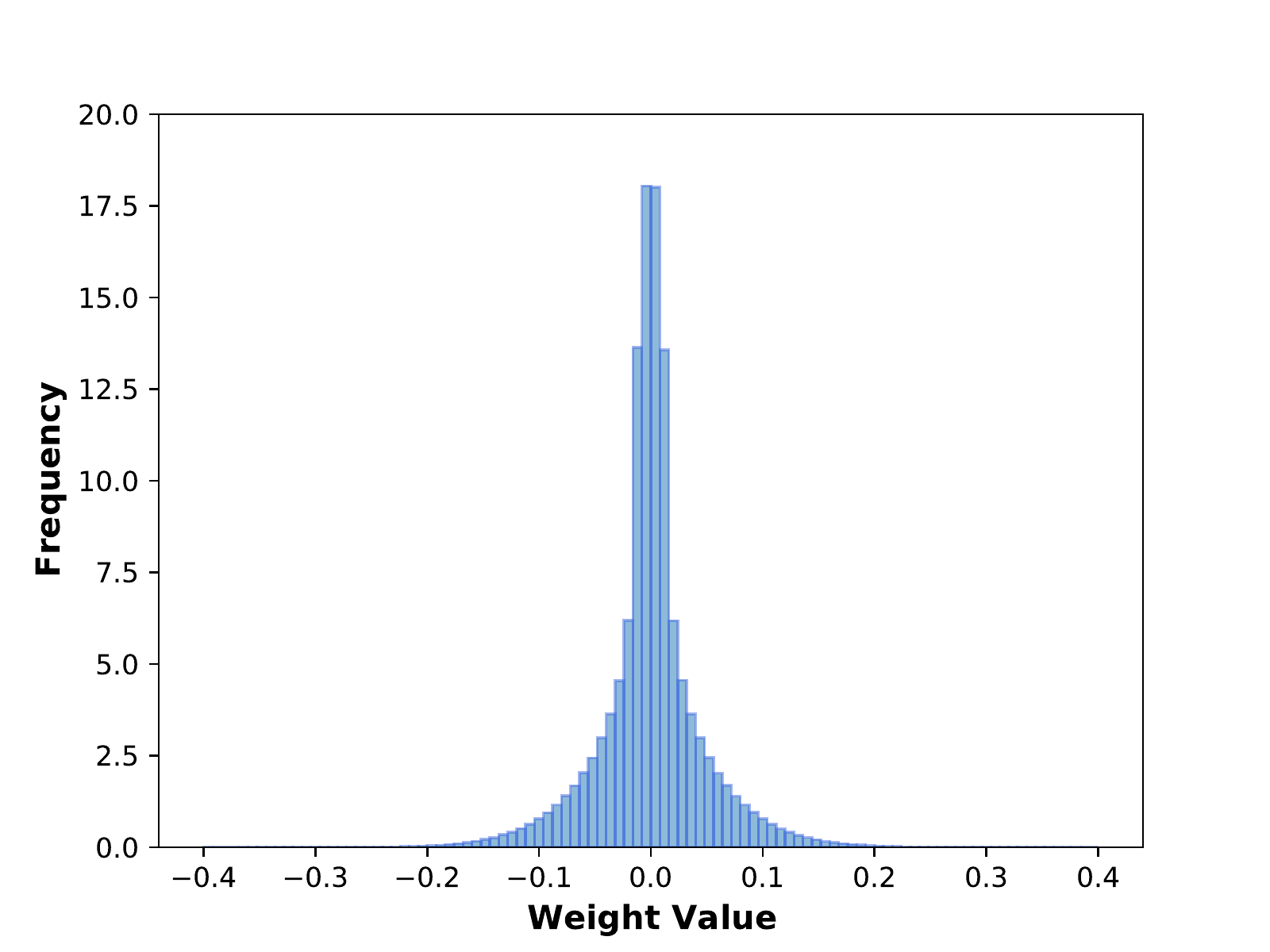}
 \label{fig:coatnetsparse-c}
}
\subfigure[Feedforward Weights]{
\includegraphics[width=0.23\linewidth]{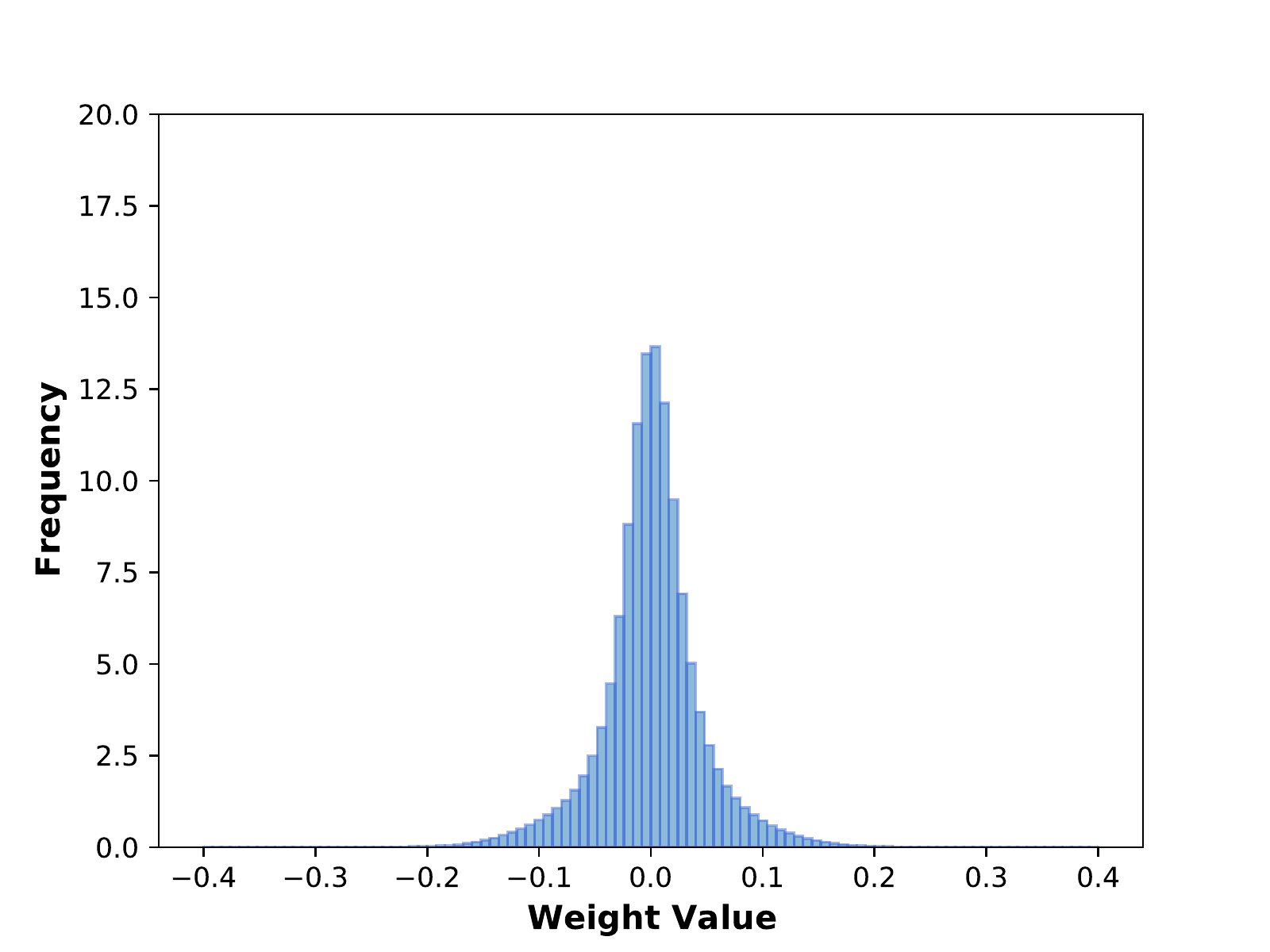}
 \label{fig:coatnetsparse-d}
}
\vspace{-0.1in}
\caption{Weight distribution for different architectural ingredients in adversarially trained CoAtNet. (a): whole weights in CoAtNet, (b): weights in convolution blocks, (c): QKV weights in attention layers, and (d): weights in feedforward blocks. The horizontal axis represents the value of model weights, while the vertical axis represents the frequency (density) of the corresponding value. } 
\label{fig:coatnetsparse}
\vspace{-0.1in}
\end{figure*}
\begin{theorem}
\label{th:th2}

Suppose the Transformer network function $\mathcal{F}=\{f_{\bm{w}}(x)\ : \bm{W}=(\bm{A}_1,\bm{W}_1), \parallel \bm{A}_1 \parallel_{p} \leq s_1, \parallel \bm{W}_1 \parallel_{p} \leq s_2, \parallel \bm{A}_1 \parallel_{1} \leq b_1, \parallel \bm{W}_{1,1} \parallel_{1} \leq b_2\}$. $\forall \gamma>0$, with probability at least $1-\delta$, we have $\forall f_{\bm{w}} \in \mathcal{F}$,
\begin{center}
\begin{flalign}
\label{eq5}
%\begin{aligned}
&&&\mathbb{P}_{(\bm{x},y)\sim \mathcal{S}}\{\exists \bm{\delta} \in \mathbb{B}(\epsilon),\quad\emph{s.t.}\quad \bm{y} \neq \arg \max [f_{\bm{w}}(\bm{x}+\bm{\delta})]_{\bm{y}^{'}} \}&\nonumber \\
	&&&\leq \frac{1}{n}\sum_{i=1}^{n}E_{i}+\frac{1}{\gamma}(\frac{4}{n^{3/2}}+\frac{60\log(n)\log(2d_{\max})}{n}s_1 s_2 C) &\nonumber \\&&&+	\frac{2\epsilon b_1 b_2}{\gamma \sqrt{n}}+3\sqrt{\frac{\log (2/p)}{2n}},&
%\end{aligned}
\end{flalign}    
\end{center}
\end{theorem}
\noindent where $\bm{w}_{1,k}$ denotes the $k$-th column of $\bm{W}_{1}$, $C = ((\frac{b_1}{s_1})_{2/3}+(\frac{b_2}{s_2})_{2/3})_{3/2}\parallel \bm{X} \parallel _{F}$, $E_{i}=\mathbbm{1}([f_{\bm{w}}(\bm{x}_i)]_{\bm{y}_i}+\frac{\epsilon}{2}\max_{k\in [K], z=\pm 1 \bm{P}\succeq 0, diag{\bm{P}} \leq 1}\langle z Q(\bm{w}_{1,k}, \bm{A}_1), \bm{P}\rangle)$, and  $ Q(\bm{w}_{1,k}, \bm{A}_1) =  \begin{bmatrix}  0 &0 & \bm{1}^{\top}\bm{A}_1^{\top} \bm{b} \\ 0 & 0 & \bm{A}_1^{\top} \bm{b}  \\ \bm{b}^{\top}\bm{A}_1 \bm{1} &  \bm{b}^{\top}\bm{A}_1 & 0\end{bmatrix}$. At this time, there is $\bm{b} = {\rm diag} (\bm{w}_{1,k})$.

 Similar to the linear classifier in Lemma~\ref{th:th1}, if we have an $\ell_1$ norm constraint on the matrix of $\bm{A}_1$, then the generalization bound of the surrogate adversarial loss does not have an explicit dependence on the dimension $d$ and will be reduced. The proof is inspired by~\cite{yin2019rademacher} and can be found in the supplementary material. Thus, we can draw the following remark.

\textbf{Remark 1.} \emph{The purpose of imposing the $\ell_1$ constraint is to increase the sparsity of the weight matrix $\bm{A}_1$, and we can conclude that the better generalization of Transformers compared with CNNs may come from the sparseness of transformer's weights caused by attention module $\bm{A}_1$.}

To further verify the reason for the high sparsity of Transformers, we visualize the weight distribution for different components in CoAtNet (a hybrid Transformer network consisting of convolution and attention), which is shown in Figure \ref{fig:coatnetsparse}. We can observe that QKV weights (main weights in the attention layer) have the highest sparsity while the weights of convolution have the lowest sparsity. Hence it is demonstrated that the attention layer could contribute to the sparsity of Transformers, which in turn improves its adversarially robust generalization. These results are consistent with Theorem \ref{th:th2} and the Remark \textcolor{red}{1}.

\section{Analyses and Discussions}
%In this section, we give more detailed analyses to help understand our above conclusions and provide possible pathways for robust architectures.

\subsection{More Detailed Analyses}
\label{sec:analysis}
\textbf{Weight loss landscape analysis.} Previous works \cite{keskar2016large, kleinberg2018alternative, jastrzkebski2018relation} have demonstrated the strong relation between model generalization ability and model weight loss landscape, and flatter loss landscape usually leads to better generalization ability. Some recent works \cite{yamada2021adversarial, kanai2021smoothness, liu2020loss, wu2020adversarial, chen2021vision} further studied this relationship under robust generalization. For example, \cite{yamada2021adversarial, liu2020loss, wu2020adversarial} have shown that adversarial training would sharpen the weight loss landscape; \cite{chen2021vision} demonstrated that ViT and MLP-Mixer converge to sharper regions than ResNet under standard training. By contrast, we investigate the weight loss landscape for different model architectures under standard and adversarial training to help better understand adversarially robust generalization. Specifically, we visualize the weight loss landscape for 2 representative architectures (ResNet and PVTv2) under standard training, and PGD-$\ell_{\infty}$ adversarial training ($\epsilon=8/255$) leveraging the filter normalization method \cite{li2018visualizing}. 

\begin{figure*}[tb]
\vspace{-0.2in}
\subfigure[ResNet vanilla training]{
\includegraphics[width=0.23\linewidth]{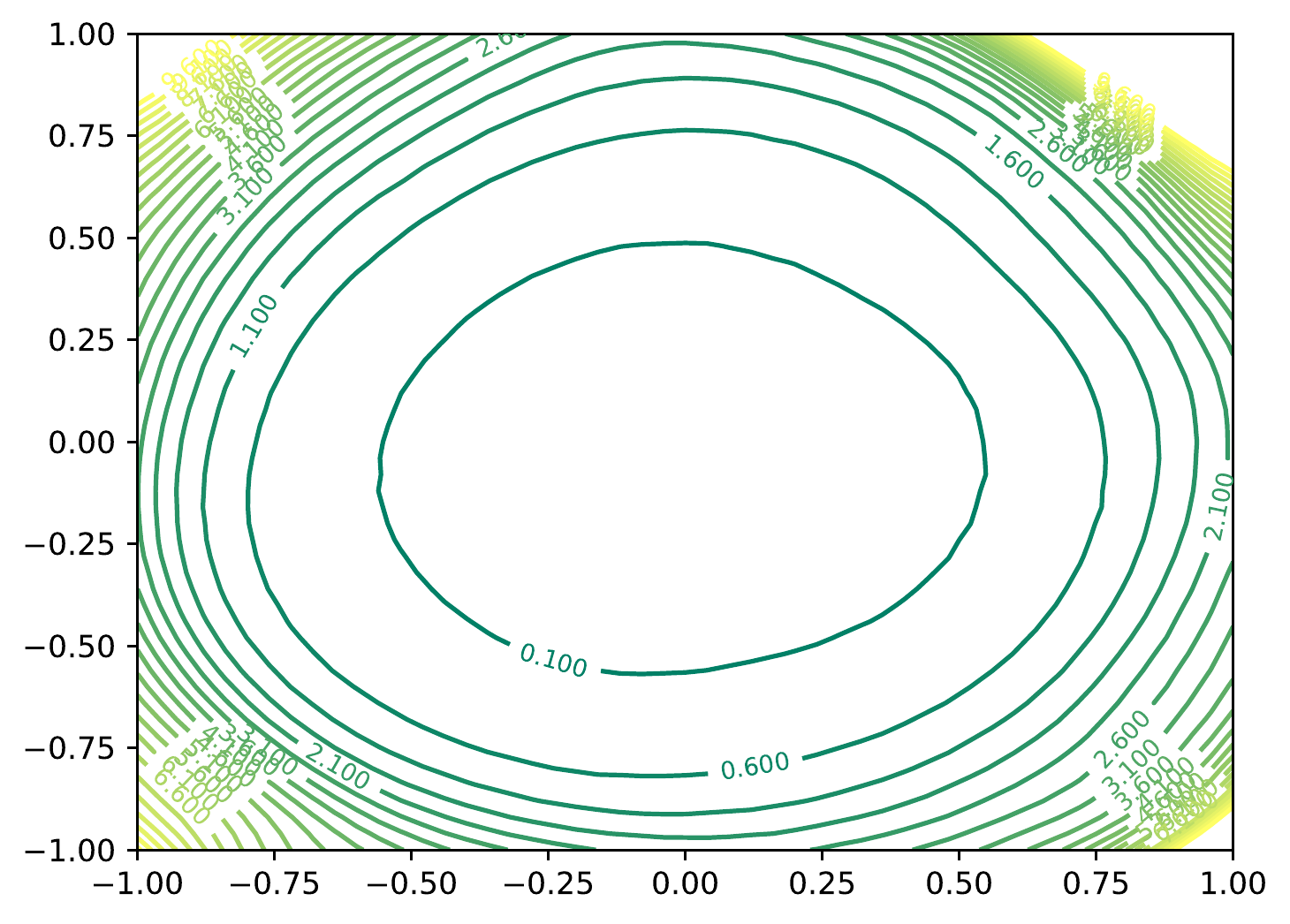}
 \label{fig:weight-a}
}
\subfigure[ResNet adv training]{
\includegraphics[width=0.23\linewidth]{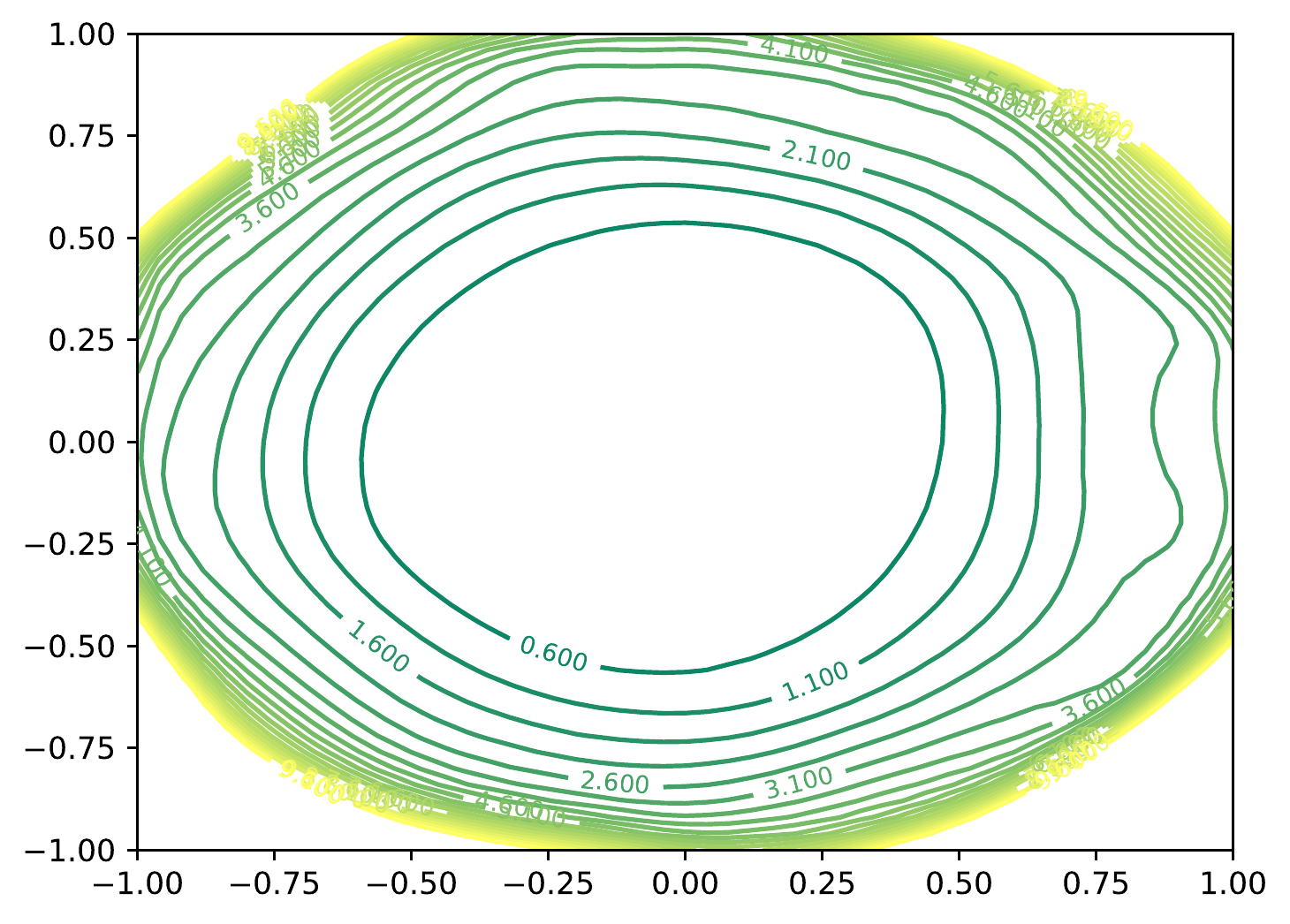}
 \label{fig:weight-b}
}
\subfigure[PVTv2 vanilla training]{
\includegraphics[width=0.23\linewidth]{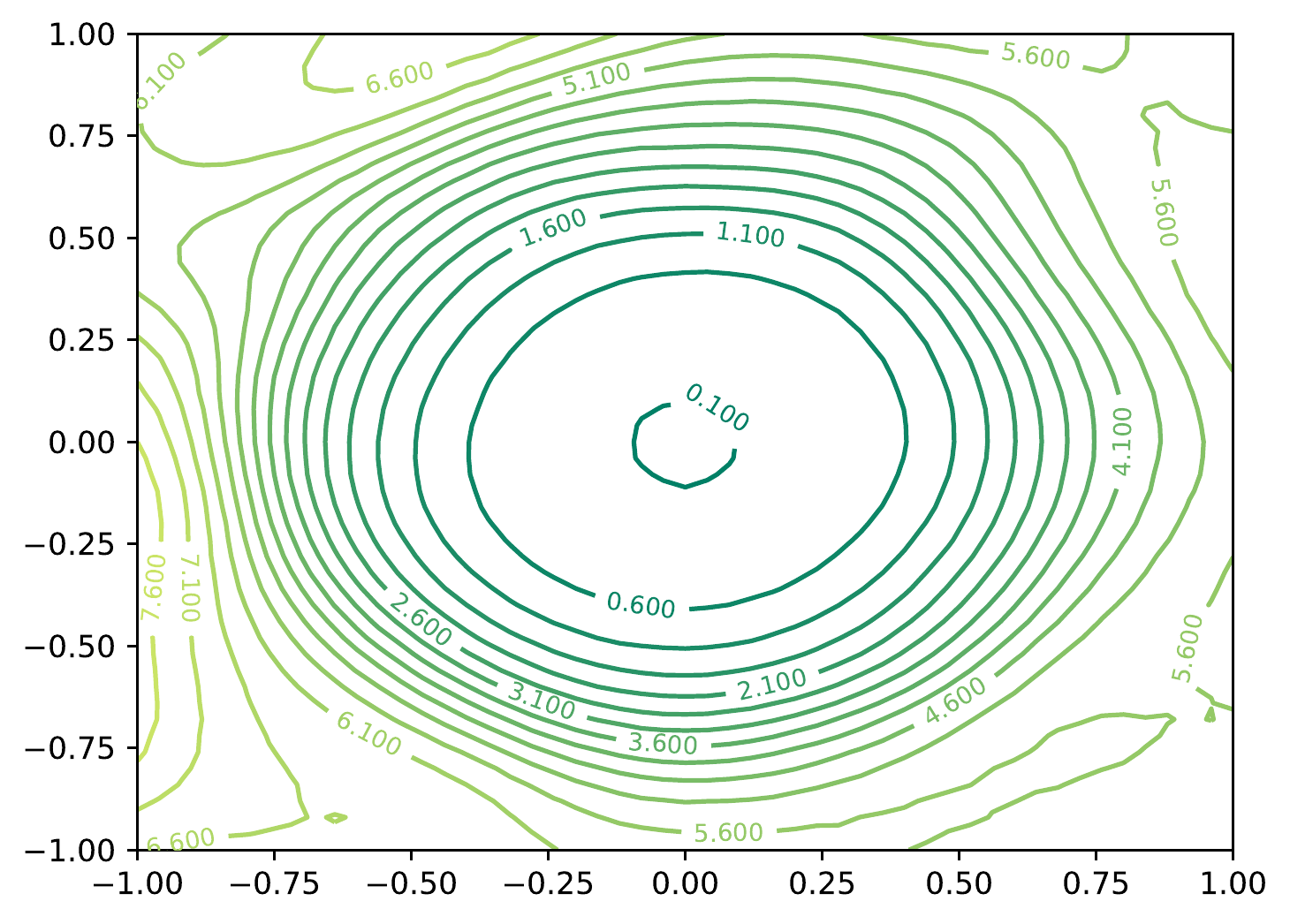}
 \label{fig:weight-e}
}
%\hspace{0.0001in}
\subfigure[PVTv2 adv training]{
\includegraphics[width=0.23\linewidth]{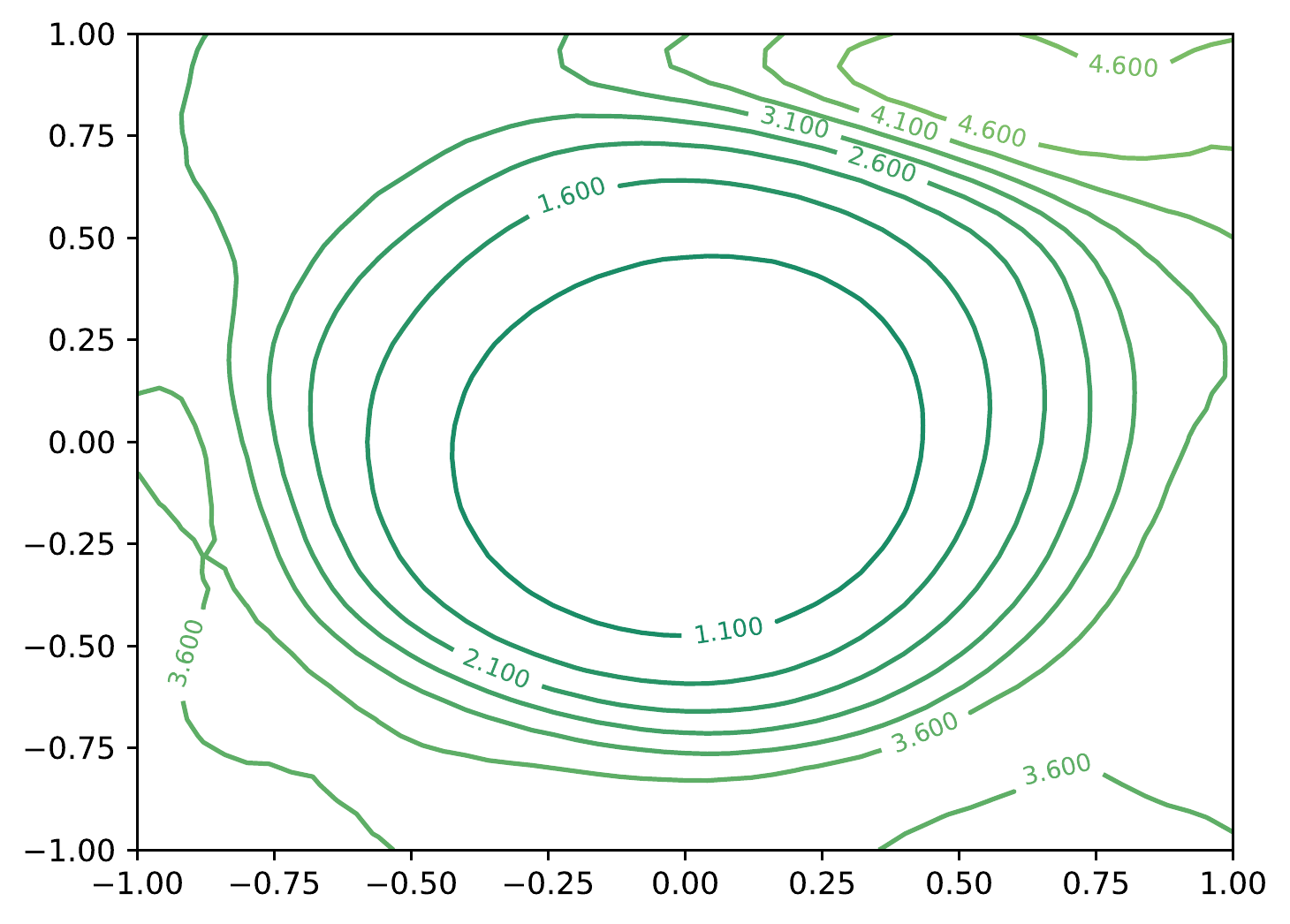}
 \label{fig:weight-f}
}

\vspace{-0.1in}
\caption{The weight loss landscape visualization for ResNet and PVTv2. The horizontal and vertical axes represent the neighborhood around the trained weights, and the contour means all model weights that have the same corresponding cross-entropy loss.} 
\label{fig:losslandscape}
\vspace{-0.1in}
\end{figure*}

As shown in Figure \ref{fig:losslandscape}, we could draw several observations: \ding{182} the loss landscapes for PVTv2 under standard training are sharper than ResNet, which is consistent with \cite{chen2021vision}; \ding{183} adversarial training make the loss landscape for ResNet sharper, which is consistent to \cite{yamada2021adversarial, liu2020loss, wu2020adversarial}; \ding{184} \emph{however, different to ResNet, the loss landscapes for PVTv2 become flattered after adversarial training.} To a certain extent, these visualization results could explain our adversarially robust generalization results in Section \ref{sec:4_1}, where PVTv2 shows comparatively better adversarially robust generalization than ResNet.

\textbf{Observations on ImageNet.} We further explore whether the tendency between Transformer and CNNs on adversarially robust generalization still exists on large-scale dataset ImageNet. In particular, we choose the representative Transformer ViT and typical CNN ResNet for experiments on ImageNet. For fair comparisons, we trained both models using similar settings as Section \ref{sec:setting}.

On ImageNet, the Worst-Case Acc of ViT is 24.77 still better than 20.08 in the ResNet, which indicates a better adversarially robust generalization of Transformers. However, the gap in the generalization ability between them shrinks as the training dataset size increases. We further analyze the weight sparsity of these adversarially-trained ViT and ResNet models, where we found that from CIFAR-10 to ImageNet the sparsity of ViT parameters improves from 0.65 to 0.95, and the sparsity of ResNet increases from 0.15 to 0.92. Recall our analysis in Equation~\ref{th:th2} where $n$ denotes the number of samples, and increasing sample numbers would decrease the minimum upper bound of the Rademacher complexity. Thus, we can conclude that the adversarially robust generalization of the model is also positively correlated with the dataset size. The generalization ability of both ViT and ResNet is generally improved by increasing the size of datasets in the same experimental setup. We speculate that the diversity of the dataset increases the sparsity of the model, and the data volume and
richness of CIFAR-10 are far less than that of ImageNet. \emph{More results can be found in the Supplementary Material.} %We will study this in the future.

\subsection{Potential Pathways}
In this part, we further provide some potential pathways for robust architectures in the future. Unless otherwise specified, the experiments in this part adopt the same experimental settings from Section \ref{sec:setting}.

\textbf{Imposing $\ell_{1}$ regularization for sparsity.} In Section \ref{sec:4_2}, we revealed that weight sparsity has a close relation with adversarially robust generalization. We then explore: \emph{could we improve adversarially robust generalization by explicitly improving the sparsity via weight regularization (e.g., $\ell_{1}$ regularization) during training?} 

\begin{figure}[tb]
\centering
\vspace{-0.1in}
\subfigure[Worst-case]{
\includegraphics[width=0.46\linewidth]{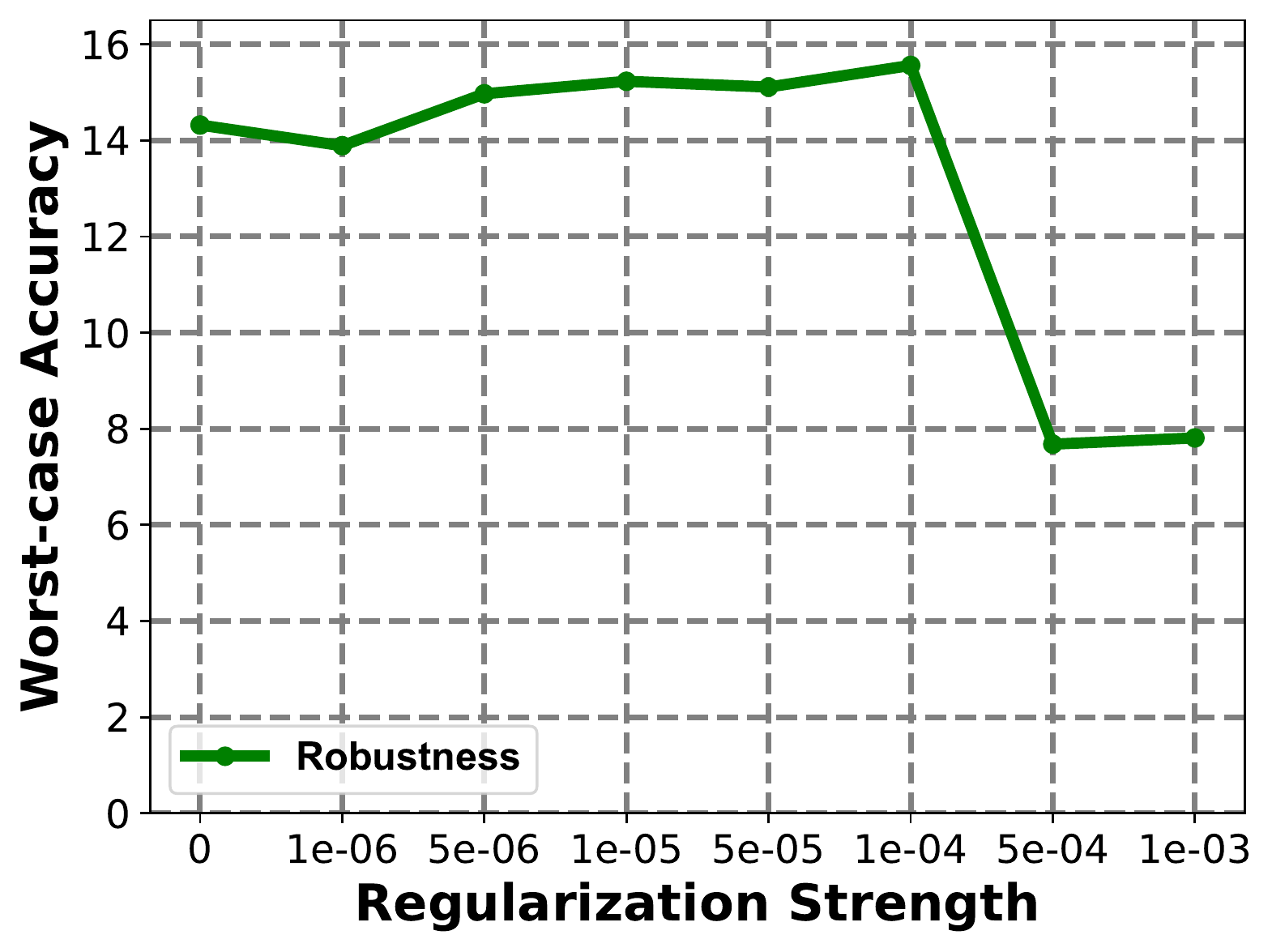}
 \label{fig:weight-b}
}
\subfigure[Clean]{
\includegraphics[width=0.46\linewidth]{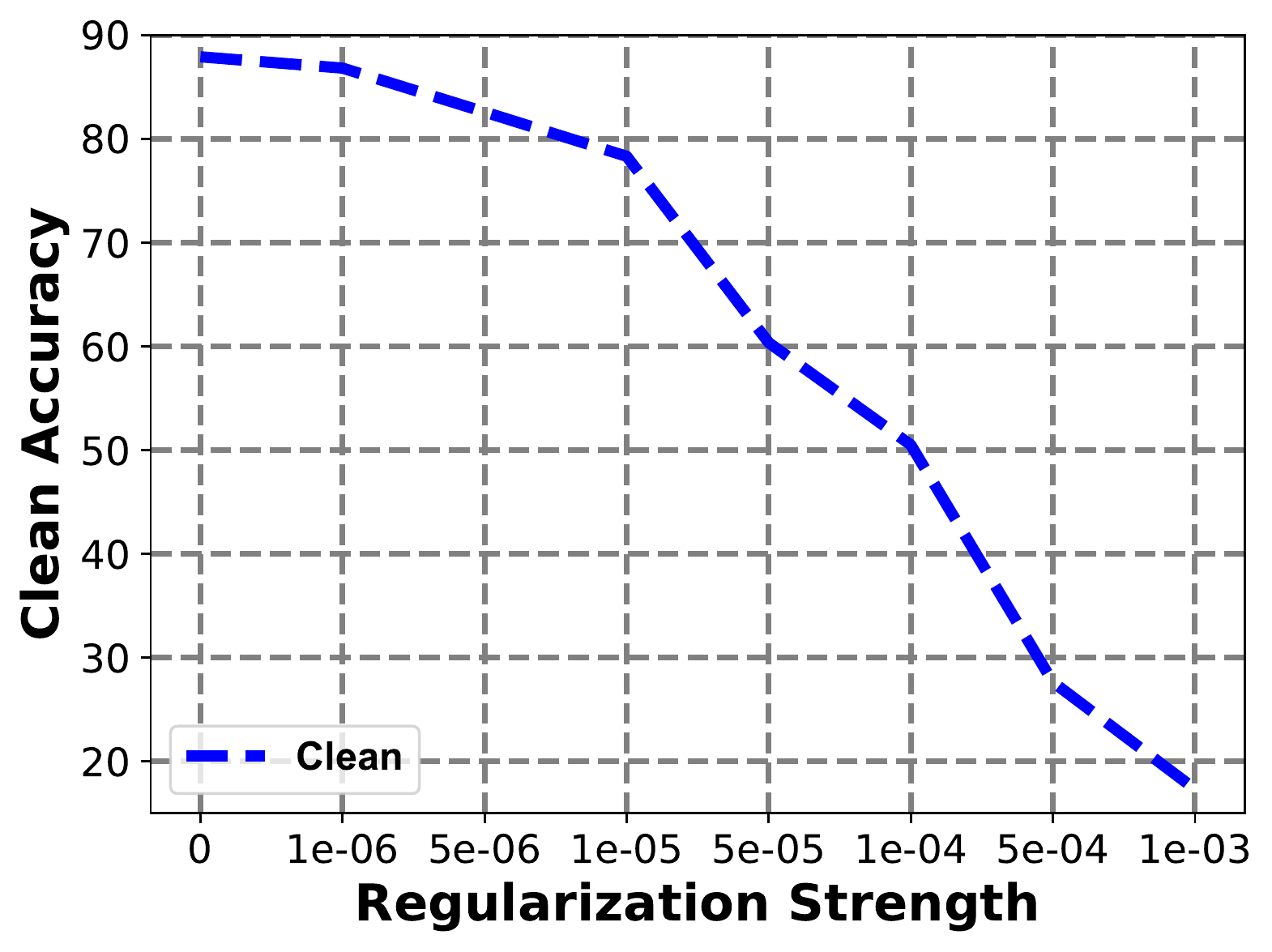}
 \label{fig:weight-b}
}
\caption{Adversarial training on ResNet with different $\ell_{1}$ regularization strengths to improve sparsity.}
\label{fig:L1_reg}
\end{figure}

To answer this question, we choose ResNet on CIFAR-10 and impose $\ell_{1}$ regularization during adversarial training with different regularization strengths. From Figure \ref{fig:L1_reg}, we can observe that imposing $\ell_{1}$ regularization can indeed improve the adversarially robust generalization. However, the improvement on adversarially robust generalization is comparatively small (\ie, +2\%) considering that ResNet may fall behind some Transformers for 10-20\% as shown in Table \ref{tab:CifarRobustGen}. Moreover, we could also notice the following obstacles: \ding{182} with the increasing of $\ell_{1}$ regularization strength, model clean accuracy drops, which indicates that there may exist a trade-off between clean example accuracy and adversarially robust generalization; and \ding{183} robust accuracy is highly sensitive to the $\ell_{1}$ regularization hyper-parameters, which may entail the careful adoption in practice considering the different model architectures and sizes.

Moreover, we also conduct experiments on directly changing the weight sparsity of Transformers during training, which shows a similar tendency (\emph{Supplementary Material}). The above analyses imply that \emph{to improve adversarially robust generalization, it would be better to introduce weight sparsity by designing architectural ingredients (e.g., attention layers) instead of adding training regularization.}

\textbf{Hybrid architecture with increased sparsity.} Since attention layers contribute sparsity to ViTs, we investigate the influence of different attention layer numbers on adversarially robust generalization. In particular, we adopt CoAtNet architecture, which is a hybrid architecture containing four network stages and each stage could be set as either the attention layer (denote $\texttt{T}$) or convolution layer (denote $\texttt{C}$). We choose 4 settings and evaluate their adversarially robust generalization on CIFAR-10 \footnote{Similar to \cite{dai2021coatnet}, we do not consider the first stage to be the attention layer due to the huge memory consumption.}. According to Table \ref{tab:convAttnLayer}, we highlight an interesting observation that the increasing attention layer improves weight sparsity, which might be beneficial to improving adversarially robust generalization.


\textbf{Patch size as receptive fields.} Patch splitting and embedding is an important procedure for ViTs, we, therefore, investigate the patch size on adversarially robust generalization. Specifically, we select 2 representative Transformers (ViT and PVTv2) and conduct evaluations on different patch sizes after adversarial training on CIFAR-10. As shown in Table \ref{tab:patchsize}, we can observe that with the decreasing of patch size, ViT and PVTv2 both achieve better accuracy on clean examples and PGD-$\ell_{\infty}$ adversarial examples (the perturbations that are trained on), while tend to show worse results towards PGD-$\ell_{1}$ and PGD-$\ell_{2}$ adversarial examples.  Similar results can be found in Section \ref{sec:4_1}, where Swin Transformer (which has window attention instead of global attention) performs better on $\ell_{\infty}$ noises while worse on $\ell_{1}$ and $\ell_{2}$ compared to ViT. In other words, \emph{Transformers tend to overfit the trained adversarial attacks with the decreasing of patch size}. Intuitively, decreasing patch size would make each pixel interact with the closer pixels thus making the whole model show the locality characteristic like CNNs, which are easier to fit the texture of trained adversarial examples yet harder to generalize on other attacks.  

%\textbf{Combination of convolution and attention layers.} Unlike early transformers that only contains attention layers \cite{dosovitskiy2020image}, most of the recently proposed transformers (\eg, Swin Transformer, CVT, CeiT) have hybrid architectures consisting of both attention and convolution layers at the same time. As shown in Sec \ref{sec:4_1}, these transformers show different performance, which indicates that different combinations of convolution and attention layers may cause different adversarially robust generalization. 
\begin{table}[]
\centering
\caption{The influence of convolution-attention combinations on robust generalization. $\texttt{C}$ means convolution layer and $\texttt{T}$ means attention layer, \eg, $\texttt{CTTT}$ means the first stage is the convolution layer and the last three stages are attention layer.}
\label{tab:convAttnLayer}
\normalsize
\resizebox{1\linewidth}{!}{

\begin{tabular}{lcc|cccccc}

\toprule
\multicolumn{1}{c}{\textbf{}} & \multicolumn{1}{c}{\textbf{}} & \textbf{}            & \multicolumn{6}{c}{\textbf{PGD-$\ell_{\infty}$ Adversarial Training}}                                                                                         \\
\textbf{Architecture}         & \textbf{Params (M)}           & \textbf{Vanilla} & \textbf{Clean} & \textbf{PGD-$\ell_{\infty}$} & \textbf{AA-$\ell_{\infty}$} & \textbf{PGD-$\ell_{2}$} & \textbf{PGD-$\ell_{1}$} & \textbf{Worst-case} \\ \midrule
%CoAtNet-$\texttt{CCTT}$                  & 16.99                         & 90.73                & 77.73              & 48.27                        & 39.85                       & 33.80                   & 42.30                   & 32.17                   \\
%CoAtNet-$\texttt{CCTC}$                  & 16.65                         & 91.51                & 79.54              & 47.86                        & 39.15                       & 32.61                   & 41.20                   & 30.97                   \\
CoAtNet-$\texttt{CTTT}$                  & 17.36                         & 90.83                & 74.25              & 45.70                        & 37.27                       & 31.95                   & 40.15                   & 30.30                   \\
CoAtNet-$\texttt{CTTC}$                  & 17.02                         & 91.13                & 77.84              & 46.61                        & 37.89                       & 31.61                   & 39.88                   & 29.96                   \\
CoAtNet-$\texttt{CTCC}$                  & 16.38                         & 90.69                & 78.71              & 42.36                        & 34.56                       & 27.84                   & 37.19                   & 26.69                   \\
CoAtNet-$\texttt{CCCC}$                  & 16.02                         & 91.41                & 79.14              & 43.71                        & 35.59                       & 29.03                   & 38.68                   & 27.64                   \\
%CoAtNet-$\texttt{CCCT}$                  & 16.36                         & 91.76                & 79.50              & 43.17                        & 35.40                       & 28.82                   & 37.86                   & 27.46                   \\

%CoAtNet-$\texttt{CTCT}$                  & 16.73                         & 90.17                & 78.73              & 41.13                        & 33.75                       & 26.89                   & 36.11                   & 25.88                   \\ 
\bottomrule
\end{tabular}}
\end{table}

\begin{table}[]
\centering
\caption{The influence of Transformer patch size on robust generalization for ViT and PVTv2. For each architecture we choose 3 different patch size (8, 4, 2 for ViT and 4, 2, 1 for PVT). }%All models are trained using the aligned settings in Section \ref{sec:4_1}.}
\label{tab:patchsize}
\normalsize
\resizebox{1\linewidth}{!}{
\begin{tabular}{llc|cccccc}
\toprule
\multicolumn{1}{c}{\textbf{}} & \multicolumn{1}{c}{\textbf{}} & \textbf{}            & \multicolumn{6}{c}{\textbf{PGD-$\ell_{\infty}$ Adversarial Training}}                                                                                         \\
\textbf{Architecture}         & \textbf{Patch Size}           & \textbf{Vanilla} & \textbf{Clean} & \textbf{PGD-$\ell_{\infty}$} & \textbf{AA-$\ell_{\infty}$} & \textbf{PGD-$\ell_{2}$} & \textbf{PGD-$\ell_{1}$} & \textbf{Worst-case} \\ \midrule
\multirow{3}{*}{PVTv2}        & $p=4$                         & 88.34                & 75.99              & 46.48                        & 38.18                       & 35.77                   & 46.14                   & 33.54                   \\
                              & $p=2$                         & 93.03                & 83.80              & 52.34                        & 44.04                       & 32.49                   & 39.63                   & 31.16                   \\
                              & $p=1$                         & 94.60                & 87.50              & 54.59                        & 46.58                       & 23.47                   & 24.76                   & 21.10                   \\ \midrule
\multirow{3}{*}{ViT}          & $p=8$                         & 82.30                & 72.39              & 42.77                        & 35.04                       & 32.74                   & 42.61                   & 30.72                   \\
                              & $p=4$                         & 86.73                & 78.76              & 46.02                        & 38.00                       & 30.86                   & 39.27                   & 29.24                   \\
                              & $p=2$                         & 85.99                & 77.37              & 45.45                        & 37.95                       & 25.36                   & 30.15                   & 23.78                   \\ \bottomrule
\end{tabular}}
\end{table}

\textbf{Considering generalization on common corruptions.} In our main experiments, we discuss the adversarially robust generalization towards different adversarial data distributions, while the generalization of adversarial training towards common corruptions is also of great significance. Thus, for all the 20 adversarially trained architectures, we also evaluate their performance on clean examples and common corruptions using CIFAR-10-C and ImageNette-C dataset \cite{hendrycks2018benchmarking} that consists of 15 diverse types of corruptions (\eg, fog, guassian noise). 

As shown in Table \ref{tab:cleanAndNatural}, we can observe that as for generalization of adversarial training towards clean examples and corruptions on CIFAR-10, CNNs behave better than Transformers, showing a much smaller clean accuracy drop (\eg, 95.60 $\rightarrow$ 87.92 for ResNet, while 91.58 $\rightarrow$ 80.44 for Swin Transformer) and better CIFAR-10-C accuracy; while on ImageNette some Transformers (\eg, CoAtNet, ViTAE) surpass CNNs and achieve the best robust generalization on clean examples and corruptions. It is important to note that our studies differ from recent studies that primarily benchmark Transformers and CNNs robustness on ImageNet under standard training \cite{zhou2022understanding}, where we focus on adversarial training. Thus, the generalization of different adversarially-trained architectures towards common corruptions and clean examples is still ambiguous. We conjecture that \ding{182} common corruptions and adversarial examples might be drawn from different distributions with a comparatively huge gap, and it would be highly non-trivial to achieve a universally better architecture after adversarial training; and \ding{183} there exists several factors that may influence the robustness on common corruptions (\eg, datasets, input size) \cite{zhou2022understanding}.

\begin{table}[tb]

\centering
\caption{Generalization of adversarially-trained models towards clean data and common corruptions on CIFAR-10. For common corruptions, we leverage mean Corruption Error (mCE) \cite{hendrycks2018benchmarking} and use 1-mCE as our CIFAR-10-C Acc (the higher the better). Results on ImageNette are in Supplementary Material.}
\label{tab:cleanAndNatural}
\resizebox{1\linewidth}{!}{
\begin{tabular}{l|cccc}
\toprule
                                             & \multicolumn{4}{c}{\textbf{CIFAR-10 Dataset}}                                                                                                                                    \\
                                             & \multicolumn{2}{c}{\textbf{Vanilla Training}}                                                & \multicolumn{2}{c}{\textbf{PGD-$\ell_{\infty}$ Training}}                                                                                   \\
{\textbf{Architecture}} & {\textbf{Vanilla Acc}} & {\textbf{CIFAR-C Acc}} & {\textbf{Clean Acc}} & {\textbf{CIFAR-C Acc}}  \\ \midrule
WideResNet                                   & 96.47                                       & 83.91                                          & 89.54                                     & 81.48                                            \\
ResNet                                       & 95.60                                       & 81.20                                          & 87.92                                     & 79.24                                           \\
PreActResNet                                 & 95.86                                       & 82.18                                          & 87.96                                     & 78.99                                           \\
ResNeXt                                      & 95.64                                       & 80.43                                          & 87.12                                     & 77.76                                              \\
VGG                                          & 94.01                                       & 81.22                                          & 84.30                                     & 75.85                                           \\
DenseNet                                      & 94.42                                       & 79.73                                          & 83.23                                     & 74.60                                          \\
BoTNet                                       & 94.16                                       & 81.04                                          & 80.76                                     & 72.72                                          \\
CCT                                          & 92.27                                       & 78.99                                          & 81.23                                     & 72.56                                          \\

ViT                                          & 86.73                                       & 77.06                                          & 78.76                                     & 71.87                                      \\
Swin Transformer                             & 91.58                                       & 77.61                                          & 80.44                                     & 71.36                                           \\ 
CPVT                                         & 90.34                                       & 79.66                                          & 78.57                                     & 70.74                                           \\
LeViT                                        & 89.01                                       & 78.31                                          & 77.10                                     & 70.48                                              \\
CoAtNet                                      & 90.73                                       & 79.91                                          & 77.73                                     & 70.27                                           \\
MobileViT                                    & 91.47                                       & 80.48                                          & 77.52                                     & 70.15                                           \\
PVTv2                                        & 88.34                                       & 79.84                                          & 75.99                                     & 69.12                                         \\
ViTAE                                        & 88.24                                       & 75.86                                          & 75.42                                     & 67.58                                         \\
PoolFormer                                   & 89.26                                       & 77.57                                          & 73.66                                     & 66.45                                              \\
CVT                                          & 87.81                                       & 75.10                                          & 73.76                                     & 66.28                                           \\ 
CeiT                                         & 85.24                                       & 73.99                                          & 71.55                                     & 65.07                                          \\
MLP-Mixer                                    & 83.43                                       & 70.70                                          & 62.86                                     & 57.09                                            \\

 \bottomrule
\end{tabular}}
\vspace{-0.2in}
\end{table}

\section{Conclusion}
%Adversarial training often suffers from huge robustness generalization gap on unseen testing adversaries, deemed as the adversarially robust generalization problem. 
We systematically examined the relationship between adversarially robust generalization and architectural design. Based on the extensive experiments, we found that Vision Transformers (\eg, PVT, CoAtNet) often yield better adversarially robust generalization while CNNs tend to overfit on specific attacks and fail to generalize on multiple adversaries. We then conduct theoretical analysis using Rademacher complexity, where we found that the higher weight sparsity contributes significantly towards the better adversarially robust generalization of Transformers, which can be often achieved by the attention layers. %We hope our paper can help to better understand the mechanism of designing robust architectures.

%\section*{Acknowledgement}

\scriptsize{\textbf{Acknowledgement.} This work was supported by the National Key Research and Development Plan of China (2021ZD0110601), the National Natural Science Foundation of China (62022009 and 62206009), and the State Key Laboratory of Software Development Environment.}

\normalsize
\bibliography{main}
\bibliographystyle{ieee}

\end{document}